\documentclass[10pt,twocolumn,letterpaper]{article}

\usepackage{cvpr}              
\usepackage{amsmath}
\DeclareMathOperator*{\argmin}{arg\,min}

\usepackage[table]{xcolor}
\usepackage{stfloats}
\definecolor{LightRed}{rgb}{1, 0.9, 0.9}  
\definecolor{LightBlue}{rgb}{0.9, 0.9, 1} 

\definecolor{cvprblue}{rgb}{0.21,0.49,0.74}
\usepackage[pagebackref,breaklinks,colorlinks,citecolor=cvprblue]{hyperref}

\def\paperID{2964} 
\def\confName{CVPR}
\def\confYear{2024}

\title{From Text to Pixels: A Context-Aware Semantic Synergy Solution for Infrared and Visible Image Fusion}

\author{
	Xingyuan Li\textsuperscript{\rm 1},
    Yang Zou\textsuperscript{\rm 2},
	Jinyuan Liu\textsuperscript{\rm 1},
    Zhiying Jiang\textsuperscript{\rm 1},
	Long Ma\textsuperscript{\rm 1},
	Xin Fan\textsuperscript{\rm 1},
	Risheng Liu\textsuperscript{\rm 1,3\thanks{Corresponding author}} \\
    \small \textsuperscript{\rm 1} School of Software Technology, Dalian University of Technology \\
    \small \textsuperscript{\rm 2} School of Computer Science, The University of Sydney \\
    \small \textsuperscript{\rm 3} Peng Cheng Laboratory \\
}

\begin{document}
\maketitle


\begin{abstract}
With the rapid progression of deep learning technologies, multi-modality image fusion has become increasingly prevalent in object detection tasks. Despite its popularity, the inherent disparities in how different sources depict scene content make fusion a challenging problem. Current fusion methodologies identify shared characteristics between the two modalities and integrate them within this shared domain using either iterative optimization or deep learning architectures, which often neglect the intricate semantic relationships between modalities, resulting in a superficial understanding of inter-modal connections and, consequently, suboptimal fusion outcomes. To address this, we introduce a text-guided multi-modality image fusion method that leverages the high-level semantics from textual descriptions to integrate semantics from infrared and visible images. This method capitalizes on the complementary characteristics of diverse modalities, bolstering both the accuracy and robustness of object detection. The codebook is utilized to enhance a streamlined and concise depiction of the fused intra- and inter-domain dynamics, fine-tuned for optimal performance in detection tasks. We present a bilevel optimization strategy that establishes a nexus between the joint problem of fusion and detection, optimizing both processes concurrently. Furthermore, we introduce the first dataset of paired infrared and visible images accompanied by text prompts, paving the way for future research. Extensive experiments on several datasets demonstrate that our method not only produces visually superior fusion results but also achieves a higher detection mAP over existing methods, achieving state-of-the-art results.
\end{abstract}


\section{Introduction}
\label{sec:intro}
\begin{figure}[!hpt]
    \centering
    \vspace{-1em}
    \setlength{\belowcaptionskip}{-10pt}
    \includegraphics[width=1\linewidth]{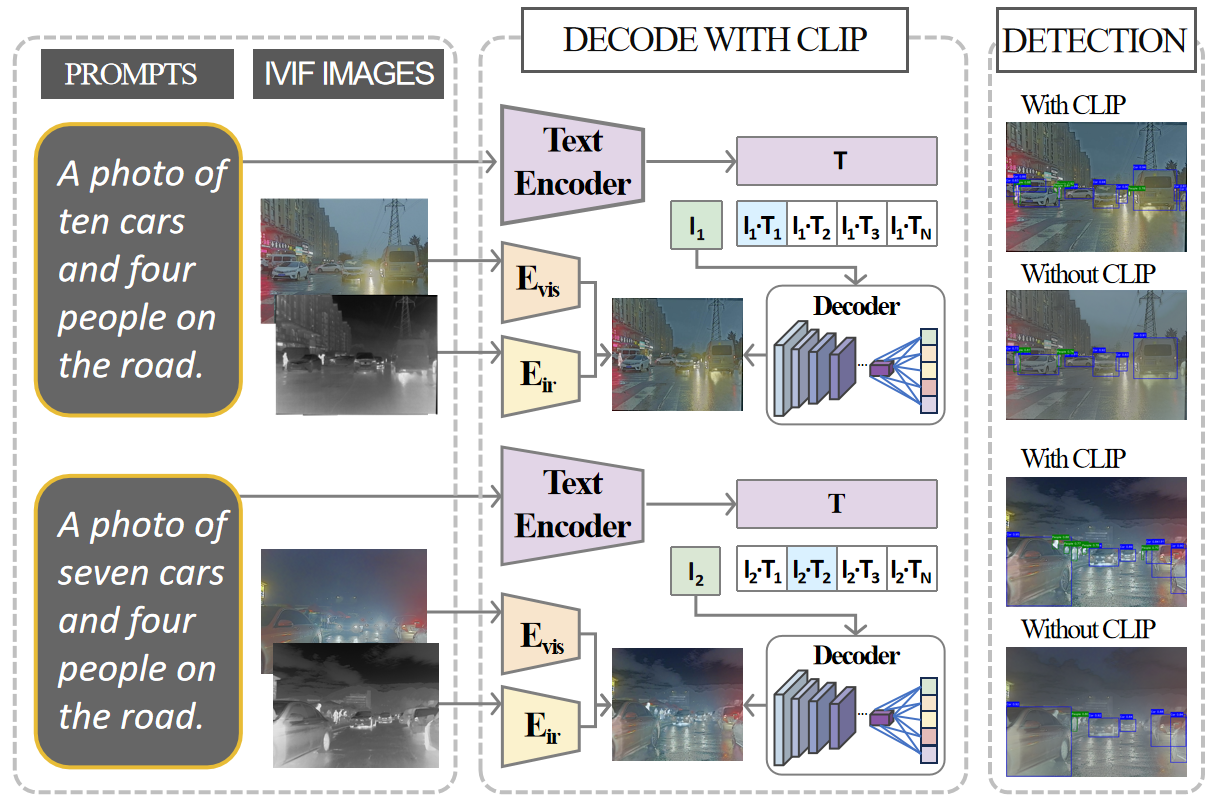}
    \caption{Schematic representation of semantic integration from textual descriptions into infrared and visible images to enhance object detection efficacy.}
    \label{pics:cover}
\end{figure}
The advent and progression of deep learning technologies have paved the way for innovative approaches in multi-modality image fusion, particularly in the realm of night vision system \cite{night_vision} and medical imaging \cite{medical_imaging}. The essence of multi-modality fusion lies in its ability to amalgamate information from diverse modalities thereby enhancing the robustness and accuracy of subsequent tasks like object detection \cite{detection}. Unfortunately, the inherent disparities and diverse representations of scene content across different modalities pose a formidable challenge, i.e, the infrared images typically exhibits diminished spatial resolution. The complexity is further amplified by factors such as image diversity, occlusions, and background interference. 

Traditional IVIF (infrared and visible image fusion) \cite{IVIF_survey} methodologies, such as spare representation \cite{sparse1, sparse2}, multi-scale transform \cite{MST1, MST2, MST3}, and subspace \cite{subspace1, subspace2} methods decompose source images into multiple hierarchical levels. Subsequently, they engage in the fusion of corresponding layers, adhering to specific, predefined rules, and reconstruct the target images in alignment with the derived fused layers. These traditional methods fuse the image in a way that heavily depends on the handcrafted feature extraction and weighting rules, fall short in addressing the intricate semantic relationships between modalities. To address this challenge, recent IVIF fusion techniques have incorporated deep learning \cite{Deep_Multi_Scale_Feature, DRF, DDcGAN}. Distinct network branches can be employed to extract features from different modalities, and multiple types of information from the same modality using different branches\cite{IVIF_survey}.

Nevertheless, either traditional or contemporary deep learning methodologies predominantly focus on enhancing fusion quality, fail to obtain the satisfied result in subsequent detection phase. A common oversight in these methods is the underestimation of the importance of modality disparities \cite{TarDAL}. For instance, multi-scale transform based techniques are anchored in predefined transforms, along with their respective decomposition and reconstruction levels. However, the lack of evaluation metrics for these transforms and levels complicates the task of discerning the intricate semantic interplay between modalities \cite{sparse_review}. This often leads to a superficial understanding of inter-modal relationships, culminating in less-than-optimal detection results.

In light of the aforementioned challenges, this paper introduces a text-guided multi-modality fusion framework, which leverages the high-level semantics derived from textual descriptions to guide the integration of semantics from infrared and visible images. Specifically, our method employs the CLIP (Contrastive Language-Image Pre-training) model \cite{CLIP} to encode high-level image semantics from text prompts, thereby facilitating a more coherent and semantically rich fusion of modalities. This not only enhances the semantic alignment between modalities but also significantly improves the model's training efficiency and performance on target tasks.

Moreover, we introduce a bilevel optimization strategy \cite{bilevel} that establishing a coherent nexus between the joint problem of fusion and detection, thereby optimizing both processes concurrently. The incorporation of codebook \cite{codebook} further refines our network's capability by discretizing the continuous feature space, thereby optimizing it for object detection tasks. Our method is particularly potent in aligning text and feature domains swiftly and enhancing performance on target tasks, thereby presenting a robust solution to the challenges posed by conventional multi-modality fusion techniques, and surpassing state-of-the-art approaches. The contributions of our work are manifold:

\begin{itemize}

\item We introduce the first text-guided multi-modality fusion perception model.

\item We employ CLIP to implement text guidance, for which we have developed the first paired IVIF detection dataset with text prompts.

\item Utilizing codebook, we enhance the generalization of the object recognition network, improve model training efficiency, and expedite the alignment of text and feature domains.

\item By employing a bilevel optimization strategy in our network, we establish a connection between fusion and detection, optimizing both tasks concurrently, achieving state-of-the-art results.

\end{itemize}

\begin{figure*}[ht!]
    \centering
    \vspace{-1em}
    \setlength{\abovecaptionskip}{2pt}
    \includegraphics[width=0.95\linewidth]{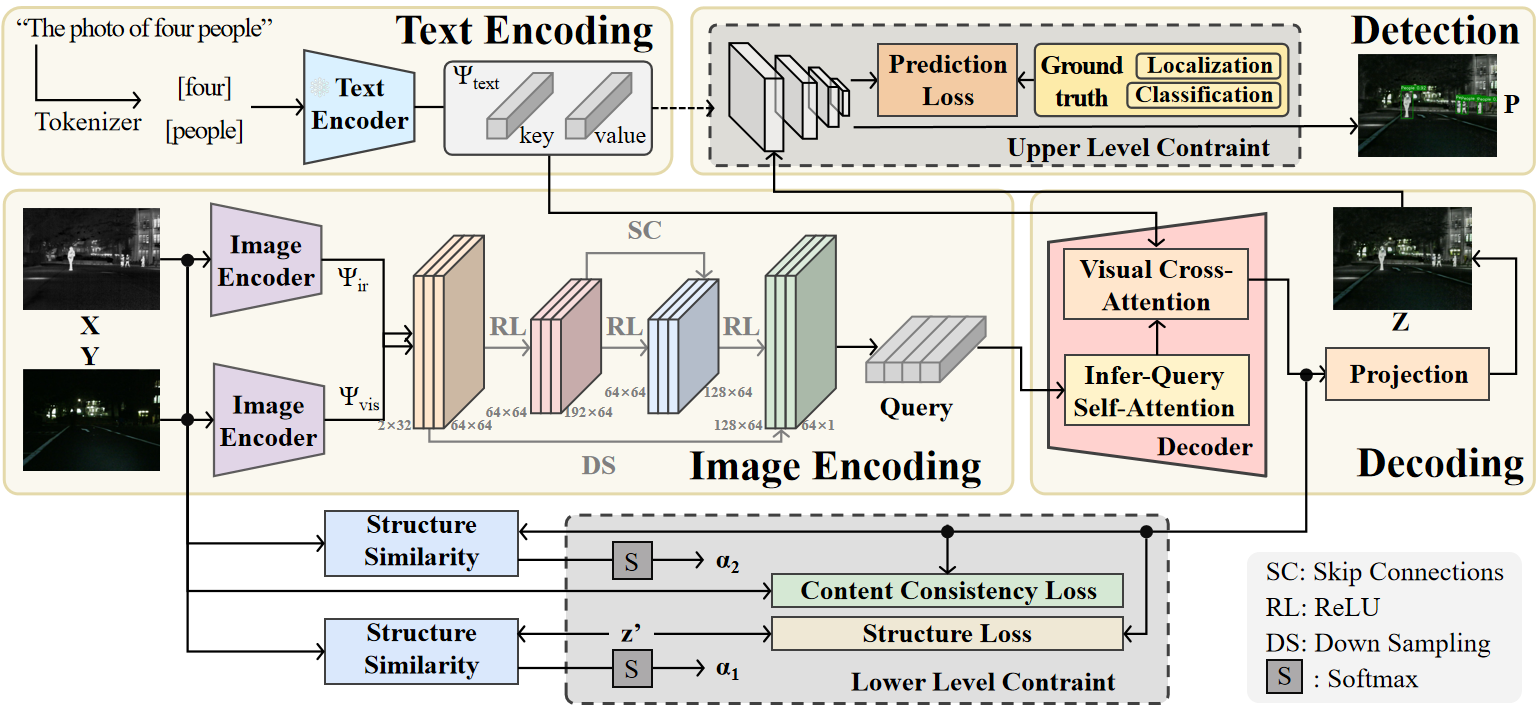}
    \caption{The overview architecture of the our proposed text-guided fusion for multi-modal image fusion and object detection.}
    \label{pics:net1}
\end{figure*}



\section{Related Works}
\label{sec:related_works}
\paragraph{CLIP Model}
Traditional pre-trained models either transform the title, description, and hashtag metadata of images into a bag-of-words multi-label classification task \cite{LWSD, LVNWD}, or explore novel model architectures and pre-training techniques \cite{VirTex, ConVIRT, ICMLM}. These approaches highlight the potential of pre-trained models to extract image representations from textual data. However, challenges such as narrow supervision in datasets like ImageNet \cite{ImageNet}, poor data efficiency, and over-reliance on fine-tuning have constrained the effectiveness of earlier models. CLIP(Contrastive Language-Image Pre-Training) \cite{CLIP}, in contrast, addresses these issues by emphasizing broader supervision, improved data utilization, and a more generalizable pre-training approach.

One of the defining features of the CLIP model is its ability to perform tasks in a zero-shot manner, eliminating the need for fine-tuning on specific datasets. As a result, CLIP has found applications in a wide array of domains. For instance, in object detection, CLIP's nuanced understanding of contextual cues in images empowers it to identify and pinpoint objects with remarkable accuracy, surpassing traditional models in challenging scenarios \cite{CLIP_detection}. Moreover, in the realm of style transfer, CLIP's inherent grasp of both content and style paves the way for generating artistically consistent and visually striking outcomes \cite{StyleCLIP}.

\paragraph{Infrared and Visible Image Fusion}
Traditional methods \cite{sparse1, sparse2, MST1, MST2, subspace1, subspace2}, such as multi-scale transform techniques \cite{MST3}, have been widely adopted due to their ability to decompose source images into multiple hierarchical levels and fuse them based on predefined rules \cite{IVIF_survey}. 

With the advent of deep learning, the convolutional neural network (CNN) is used in image fusion \cite{SMoA, IVIFDLF, LDMF}. Liu et al. \cite{MFIF} pioneered the use of convolutional neural networks (CNN) for multi-focus image fusion. However, their specifically designed network was tailored for multi-focus fusion, relying on the computation of binary maps. The DenseFuse \cite{DenseFuse} method presents a deep learning architecture for infrared and visible image fusion, which combines convolutional layers with a fusion layer and a dense block to interconnect the output of each layer.


Recently, generative adversarial networks (GAN) based IVIF fusion methods yield impressive outcomes. FusionGAN \cite{FusionGAN} has been proposed to enhance the fusion process by ensuring the generator produces images with richer details. Specifically, it adeptly retains the intensity from the infrared image while concurrently preserving the intricate details inherent in the visible image. Liu et al. \cite{TarDAL} introduced a target-aware dual adversarial learning approach, which emphasizes the importance of preserving target information during the fusion process, showcasing remarkable results in detection tasks.

\section{Method}

In this section, we delineate our proposed approach by first delve into our multi-level feature extractor. Then we elaborate the text-guided attention mechanism for feature fusion, followed by a detailed explanation of the bilevel optimization model. Finally, we propose our codebook strategy to augment our model's performance in the detection domain. 

\subsection{Multi-level Feature Extractor}
In IVIF, feature extraction is crucial for accurately representing the comprehensive features of input images. Traditional deep learning methods often rely on a fully connected layer for feature extraction, overlooking the importance of contextual information and can result in noticeable artifacts in the fused images. To counter this, our method introduces a multi-level feature extraction mechanism that captures contextual information across various scales.

As depicted in Figure~\ref{pics:net1}, the network transforms the infrared and visible images \( I \) into the feature map \( f_{in} \) through the initial convolution layer. The architecture then employs multiple convolutional paths to extract intermediate features, designed to capture information at different levels of granularity. Specifically, our model aggregates features from preceding layers through concatenation, ensuring a comprehensive representation of the input images. Our multi-level feature extractor accumulates features without relying on dilated convolutions. By using a series of convolutional layers with skip connections, our model effectively broadens its receptive field, capturing both granular and abstract details without compromising resolution. Each convolutional path in our model consistently uses a 3 × 3 kernel size. These paths, through their design, inherently possess receptive fields that offer complementary information, ensuring a richer representation of the input images.


Let \( G_{i} \) represent the feature map of the \( i^{th} \) convolution block. The output feature map \( f_{out}\) is then computed as:
\begin{equation}
f_{out} = \text{ReLU}\left( \sum_{i=1}^{6} W_{i} \odot G_{i} + b_{i} \right). 
\end{equation}

Here, \(i\) denotes the sequence number of dilated convolution paths, \(\odot\) represents element-wise multiplication, and \( W_{i} \) and \( b_{i} \) are the weights and biases for the \( i^{th} \) convolution layer, respectively. \(\text{ReLU}(\cdot) \) is the activation function.

This design ensures a diverse and comprehensive extraction of multi-modal features, preserving the structural integrity of deep features, making them well-suited for the subsequent fusion process.
\begin{figure*}[!ht]
    \centering
    \vspace{-1em}
    \setlength{\belowcaptionskip}{-10pt}
    \includegraphics[width=0.95\linewidth]{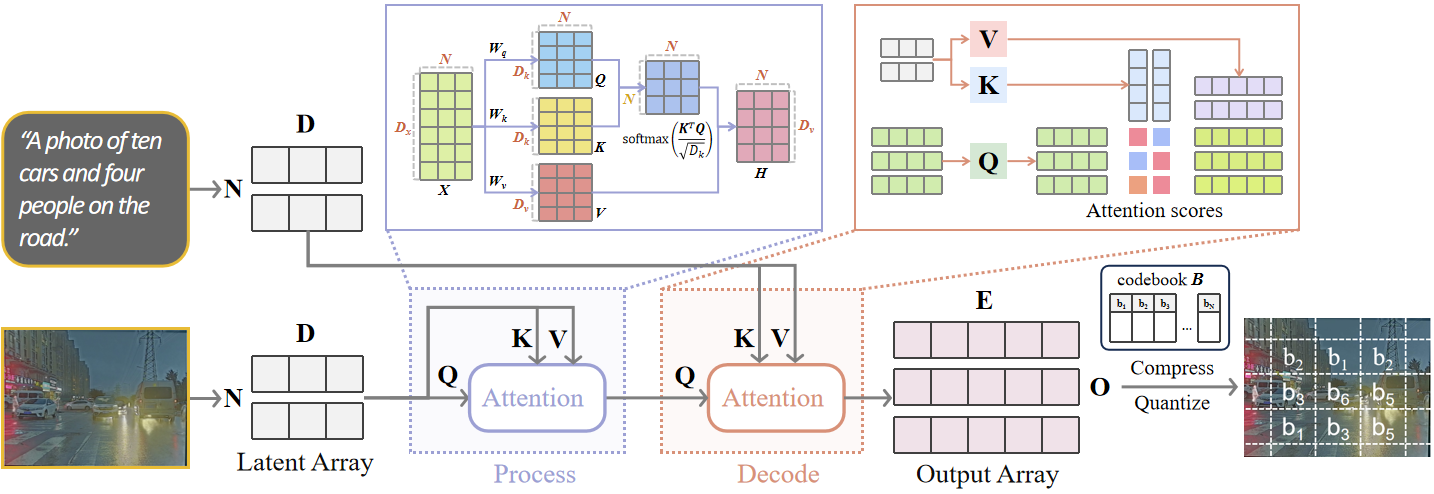}
    \caption{The procedure of our text-guided attention mechanism.}
    \label{pics:transformer}
\end{figure*}
\subsection{Text-guided Attention Feature Fusion}
After channeling the infrared and visible images into a series of intermediate features \(\mathbf{\psi_{\mathrm{image}}}\) using the multi-level feature extractor, we utilize a text-guided transformer to integrate textual semantics with image features.

To beginning with, textual descriptions are converted into a LongTensor, consisting of tokenized sequences of text prompts. This tensor serves as the input to derive the text features, \(\mathbf{\psi_{\mathrm{text}}}\), encoded by the language component of the CLIP model. To effectively incorporate the semantic information from textual descriptions into the image fusion process, we adapt \(\mathbf{\psi_{\mathrm{text}}}\) into a spatial format compatible with image features, facilitating effective interaction and fusion within a unified feature space. Then we deploy a text-guided transformer mechanism and codebook to further extract and aggregate the textual semantics with image features, as illustrated in Figure~\ref{pics:transformer}.

Firstly, we establish a self-attention-based intra-domain fusion unit to effectively integrate the global interactions within the same domain. Given the image features \(\mathbf{\psi_{\mathrm{image}}}\), the learnable weight matrices \( W_{Q} \), \( W_{K} \), and \( W_{V} \) are applied to the query \( \mathbf{Q} \), key \( \mathbf{K} \), and value \( \mathbf{V} \) matrices as:
\begin{equation}
\{\mathbf{Q}, \mathbf{K}, \mathbf{V}\} = \{\mathbf{\psi_{\mathrm{image}}}W_{Q}, \mathbf{\psi_{\mathrm{image}}}W_{K}, \mathbf{\psi_{\mathrm{image}}}W_{V}\}.
\end{equation}

Subsequently, the attention weights are computed using the dot product of the queries and keys, normalized with the softmax function. These weights are then multiplied with the values to produce the fused feature representation. The attention mechanism is defined as:
\begin{equation}
\text{Attention}\mathbf{(Q, K, V)} = \text{softmax}\left(\frac{\mathbf{Q K}^T}{\sqrt{d_k}}\right)\mathbf{V},
\end{equation}
\noindent where \( d_k \) is the dimension of the key vectors. In practice, we expand the self-attention into multi-head self-attention, allowing the attention mechanism to consider diverse attention distributions and enabling the model to capture information from multiple viewpoints. This mechanism captures global interactions within the image domain.

Following the intra-domain fusion unit, we also introduce a cross-attention-based inter-domain fusion unit to further integrate the global interactions between different domains. In this unit, the image features act as the queries, while the transformed text features \( \mathbf{\psi_{\mathrm{text}}} \) serve as the keys and values as follows:
\begin{equation}
\{\mathbf{Q}, \mathbf{K}, \mathbf{V}\} = \{\mathbf{\psi_{\mathrm{image}}}W_{Q}, \mathbf{\psi_{\mathrm{text}}}W_{K}, \mathbf{\psi_{\mathrm{text}}}W_{V}\}.
\end{equation}

This approach allows the model to weigh the image features based on textual semantics, ensuring the fused representation is influenced by the textual context. To capture diverse perspectives and ensure a comprehensive fusion of features, we employ a multi-head mechanism in the cross-attention unit. This multi-head cross-attention mechanism ensures the fused representation captures a broad spectrum of interactions between image and textual features, resulting in a robust and semantically rich feature representation suitable for downstream detection tasks.

Furthermore, we introduce a codebook-based quantization technique to refine the fused features further. The codebook maps the continuous feature space into a discrete one, representing a set of distinct feature vectors. This allows for an efficient and compact representation of the fused intra- and inter-domain interactions. Specifically, given the fused features, we compute the distances between each feature vector and all vectors in the codebook. The closest codebook vector is then chosen to represent the original feature. This process can be mathematically represented as:
\begin{equation}
\mathbf{d}(x, c) = \| x - c \|_2^2,
\end{equation}

\noindent where \(\mathbf{d}(x, c)\) denotes the distance between the feature vector \(x\) and the codebook vector \(c\). The quantized feature \(\mathbf{q}(x)\) is then defined as:
\begin{equation}
\mathbf{q}(x) = \argmin_{c \in \text{Codebook}} \mathbf{d}(x, c).
\end{equation}

By leveraging the text-guided attention feature fusion mechanism, our method not only identifies shared characteristics between the two modalities and integrates them within this shared domain but also comprehends the intricate semantic relationships between modalities. 
\begin{figure}[!ht]
    \centering
    \vspace{-1em}
    \setlength{\belowcaptionskip}{-10pt}
    \includegraphics[width=0.95\linewidth]{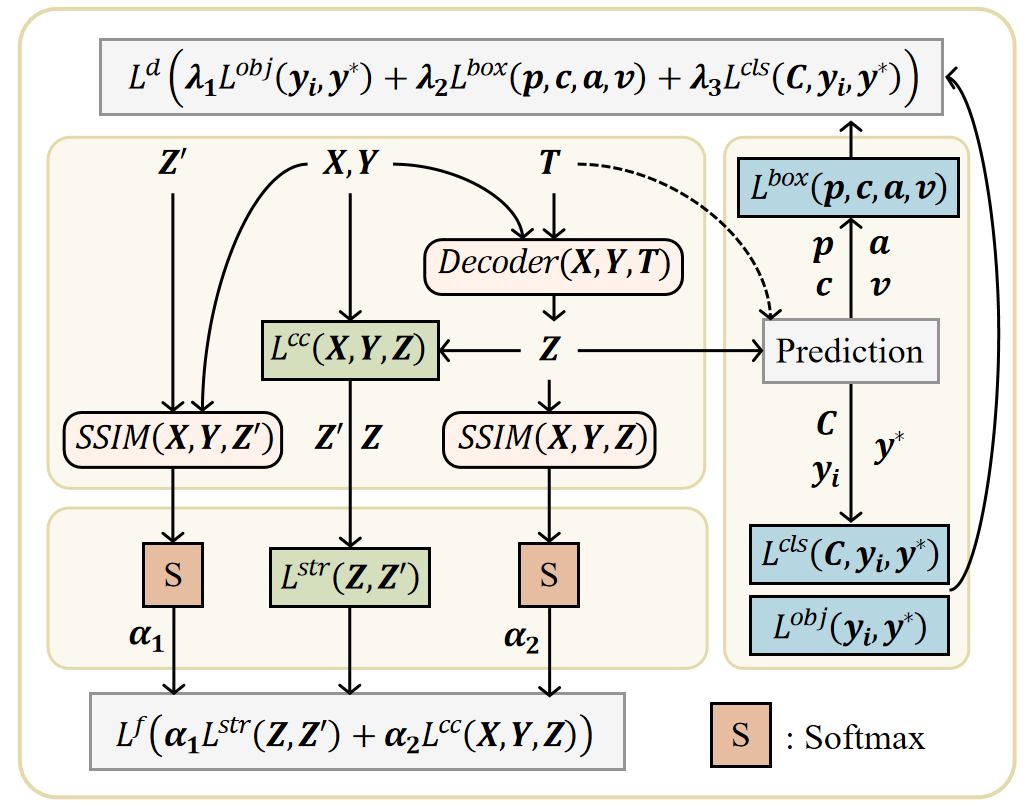}
    \caption{The framework of the bilevel optimization process.}
    \label{pics:net2}
\end{figure}

\begin{figure*}[!ht]
\setlength{\belowcaptionskip}{-10pt} 

\setlength{\intextsep}{5pt} 
    \centering
    \begin{subfigure}{.095\linewidth}
        \includegraphics[width=\linewidth, height=1.3cm]{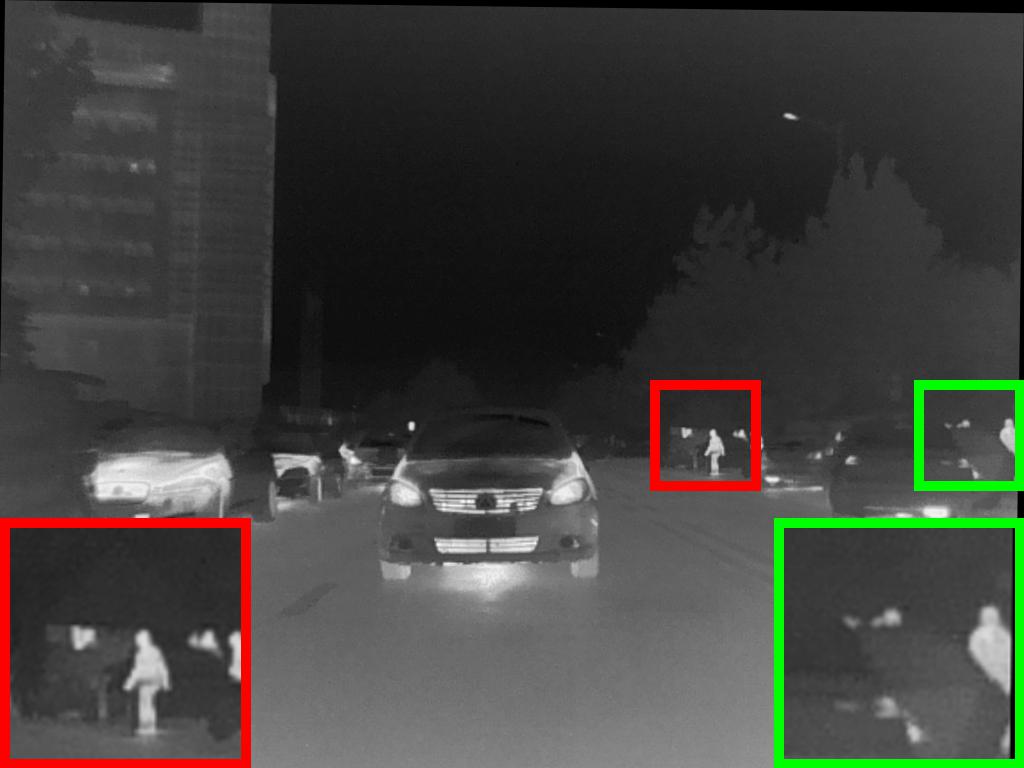}
    \end{subfigure}
    \begin{subfigure}{.095\linewidth}
        \includegraphics[width=\linewidth, height=1.3cm]{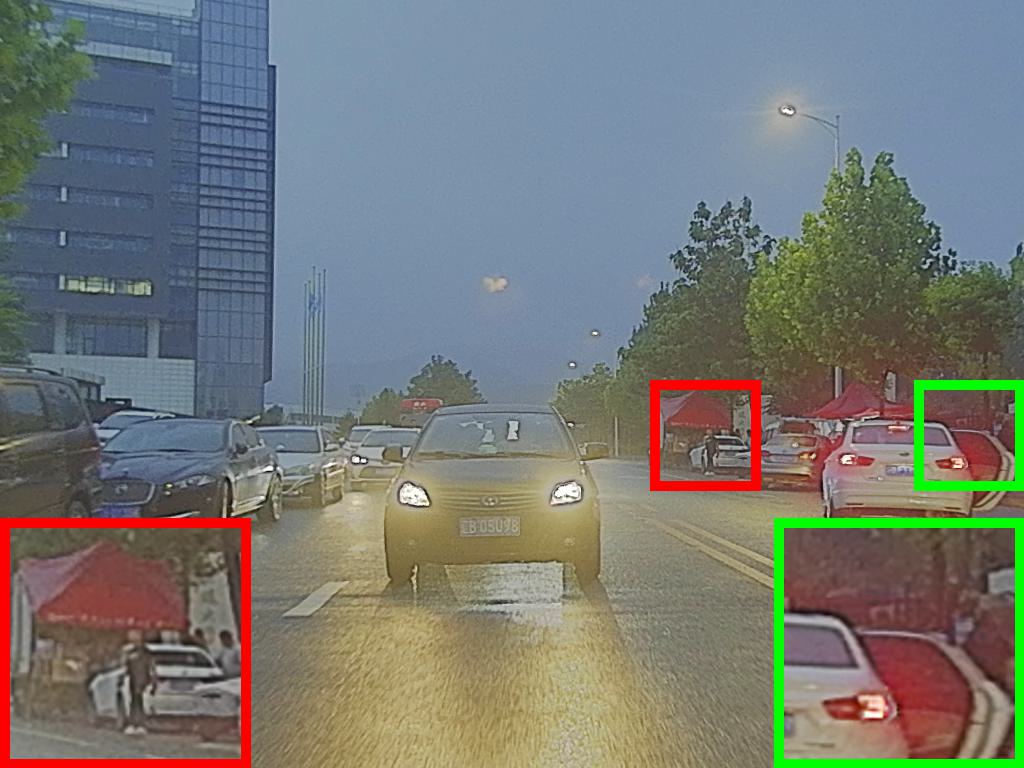}
    \end{subfigure}
    \begin{subfigure}{.095\linewidth}
        \includegraphics[width=\linewidth, height=1.3cm]{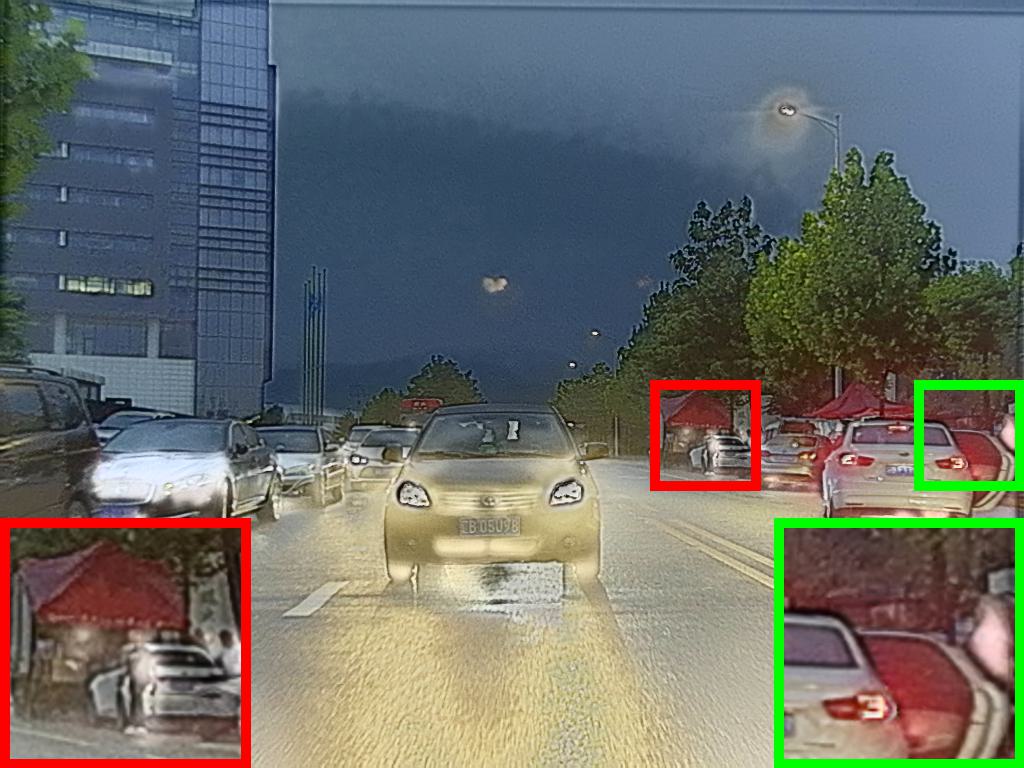}
    \end{subfigure}
    \begin{subfigure}{.095\linewidth}
        \includegraphics[width=\linewidth, height=1.3cm]{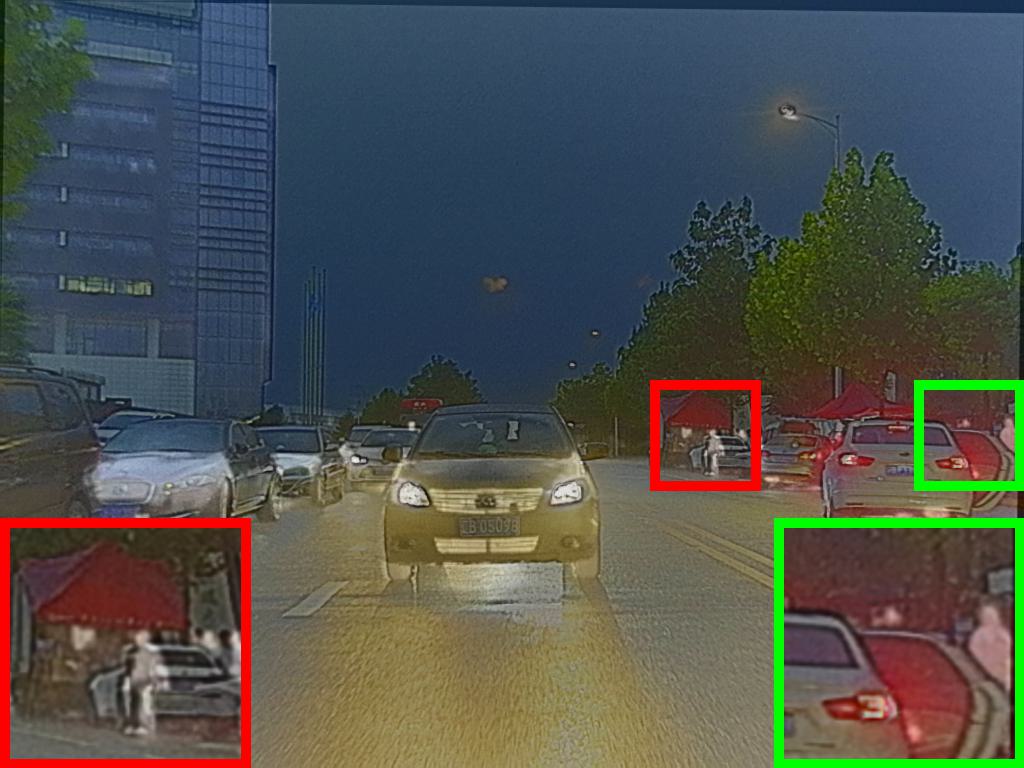}
    \end{subfigure}
    \begin{subfigure}{.095\linewidth}
        \includegraphics[width=\linewidth, height=1.3cm]{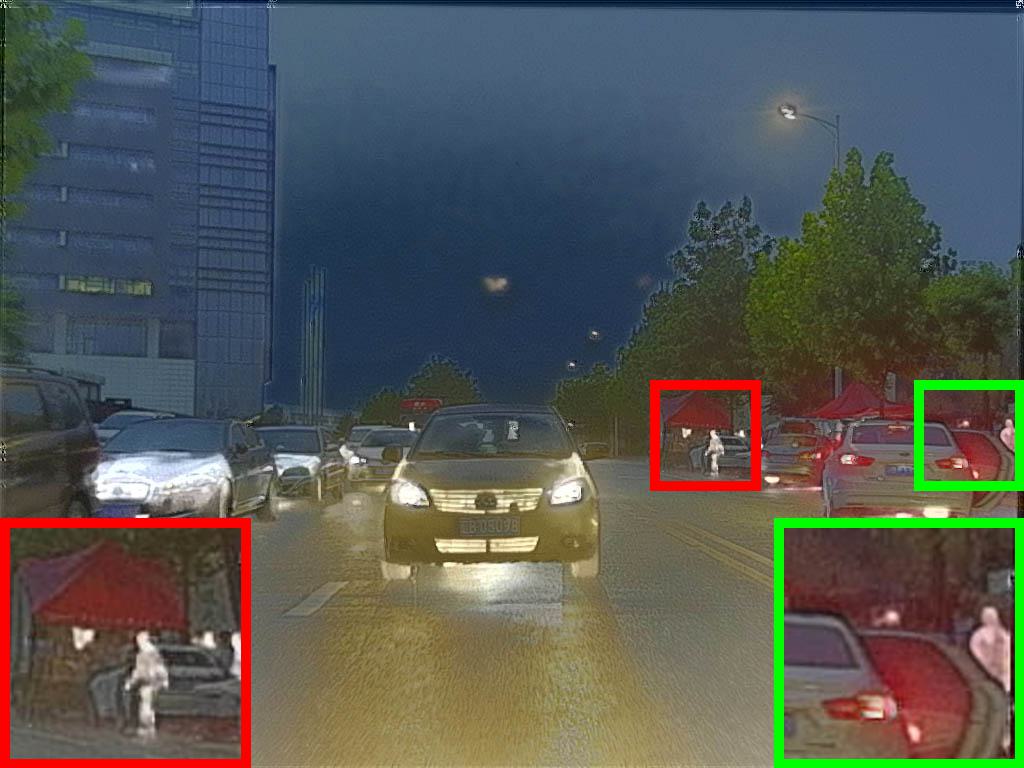}
    \end{subfigure}
    \begin{subfigure}{.095\linewidth}
        \includegraphics[width=\linewidth, height=1.3cm]{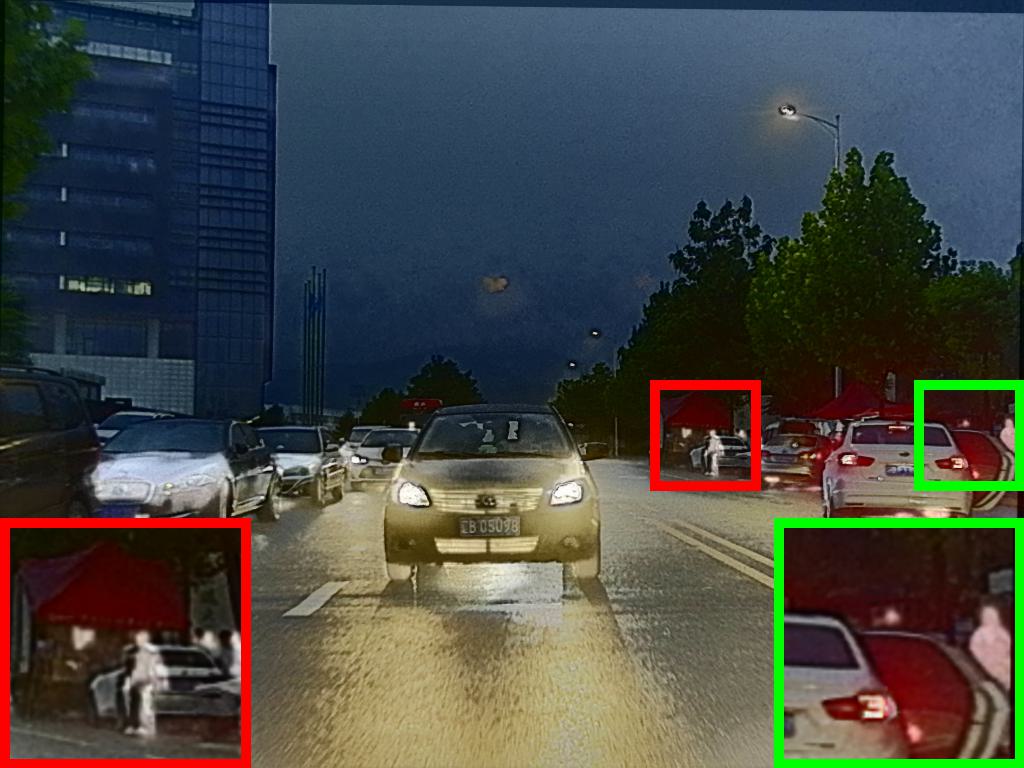}
    \end{subfigure}
    \begin{subfigure}{.095\linewidth}
        \includegraphics[width=\linewidth, height=1.3cm]{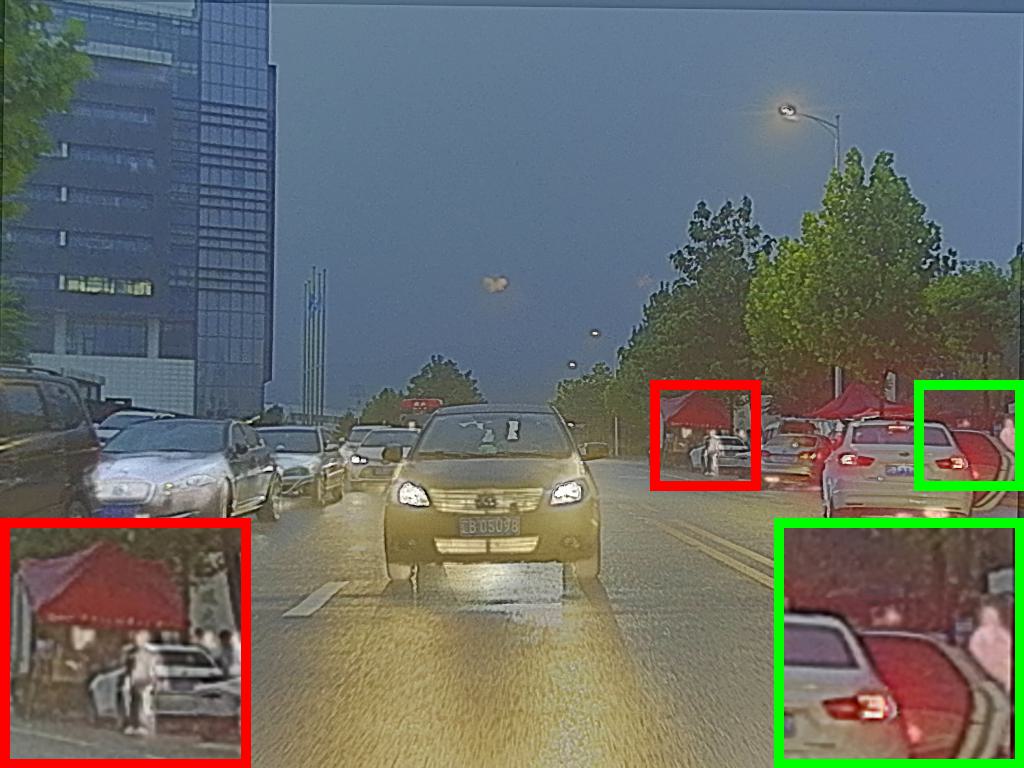}
    \end{subfigure}
    \begin{subfigure}{.095\linewidth}
        \includegraphics[width=\linewidth, height=1.3cm]{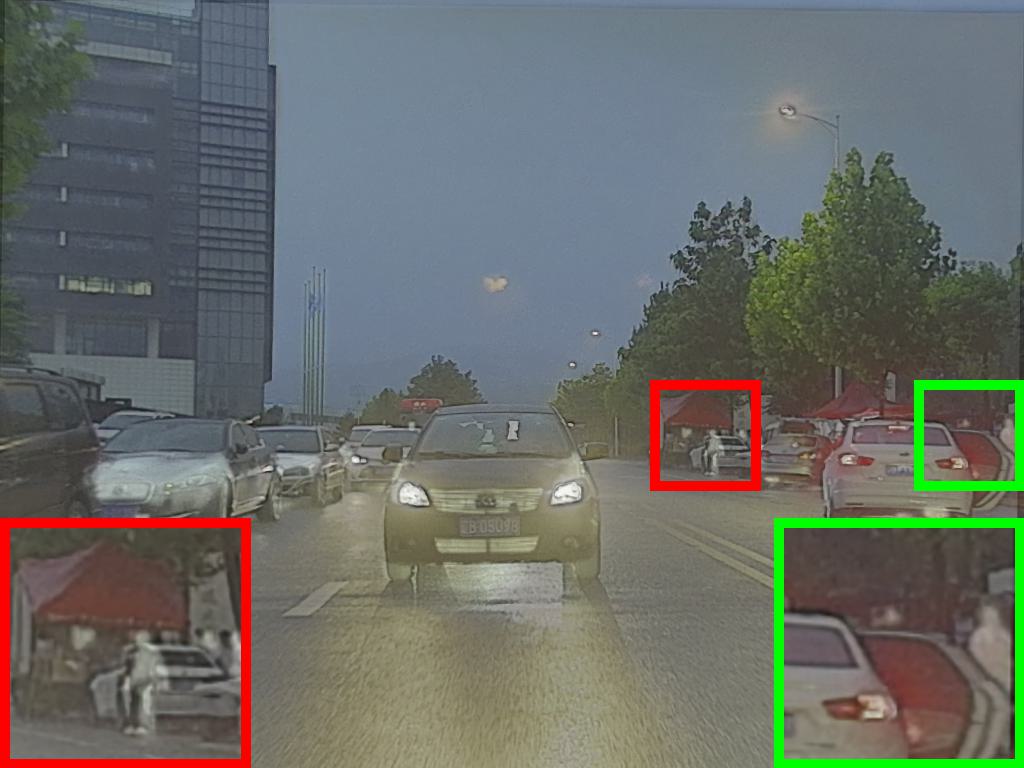}
    \end{subfigure}
    \begin{subfigure}{.095\linewidth}
        \includegraphics[width=\linewidth, height=1.3cm]{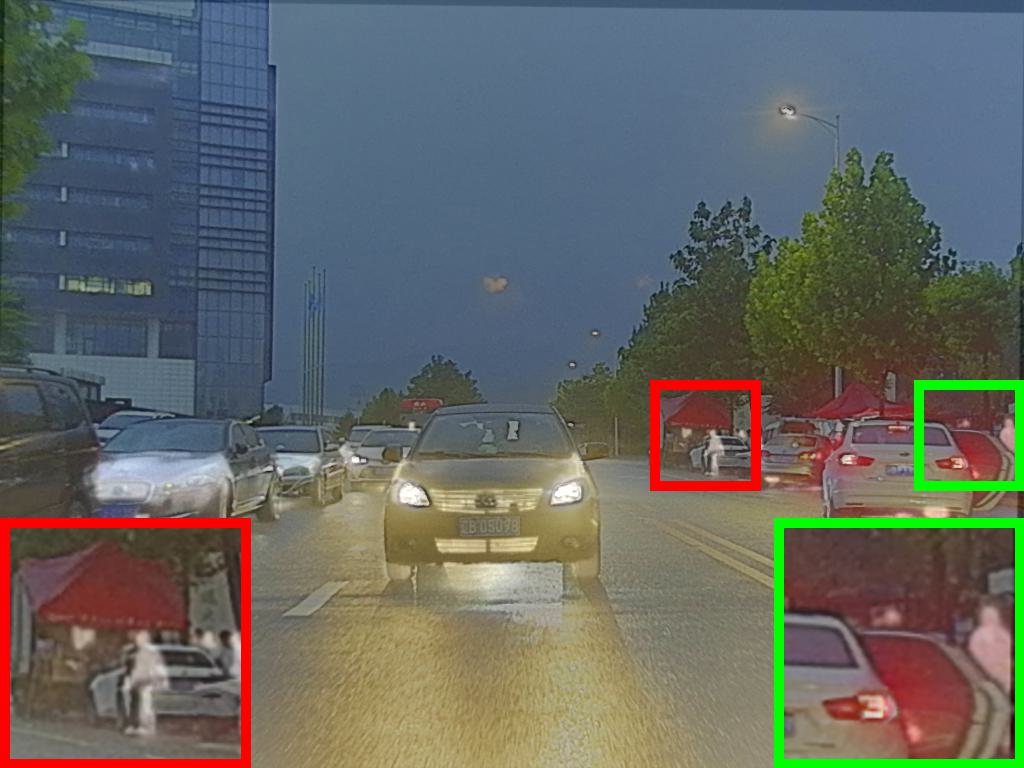}
    \end{subfigure}
    \begin{subfigure}{.095\linewidth}
        \includegraphics[width=\linewidth, height=1.3cm]{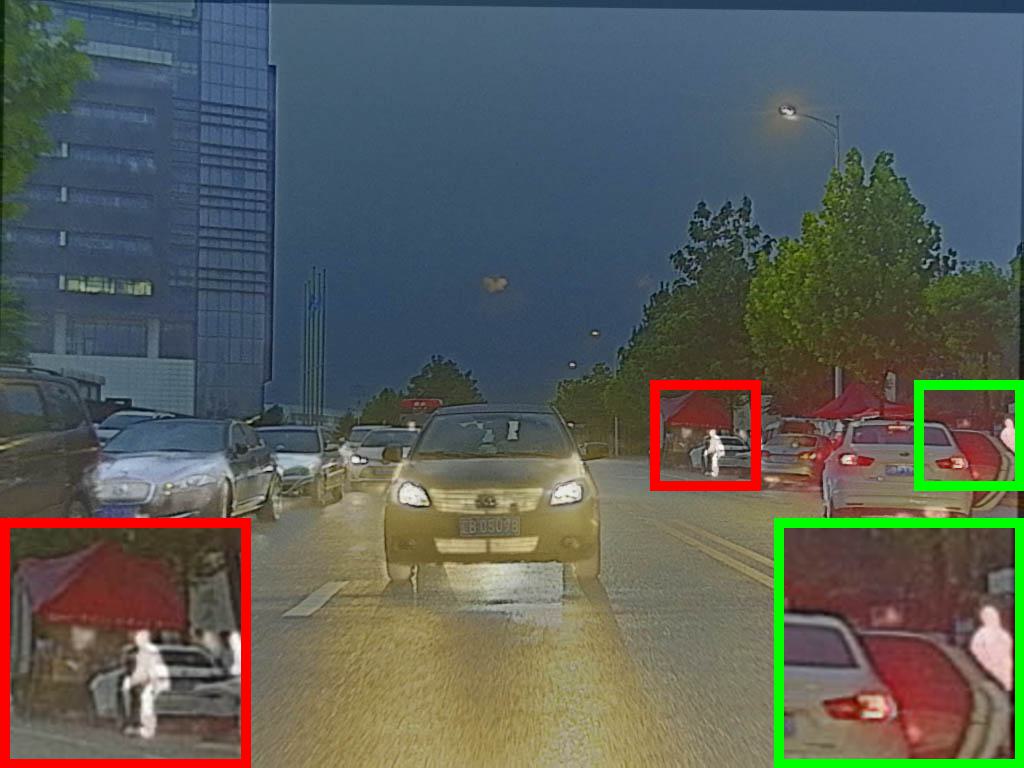}
    \end{subfigure}

    \begin{subfigure}{.095\linewidth}
        \includegraphics[width=\linewidth, height=1.3cm]{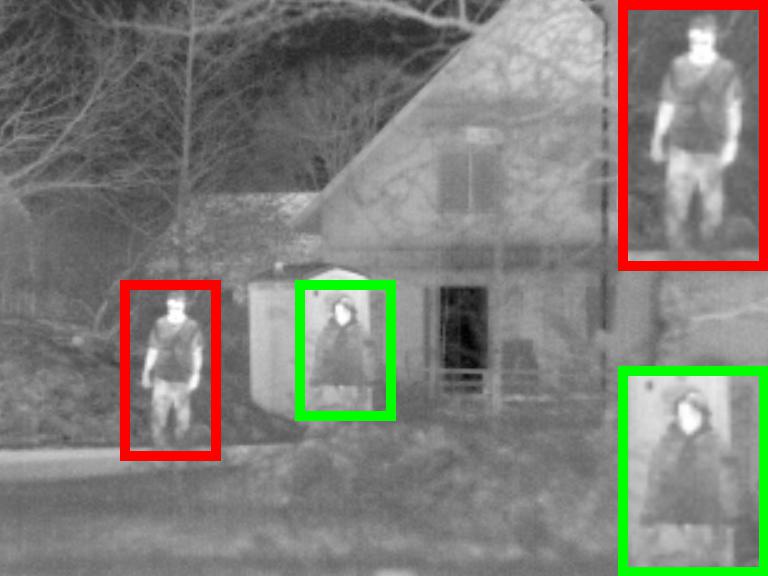}
    \end{subfigure}
    \begin{subfigure}{.095\linewidth}
        \includegraphics[width=\linewidth, height=1.3cm]{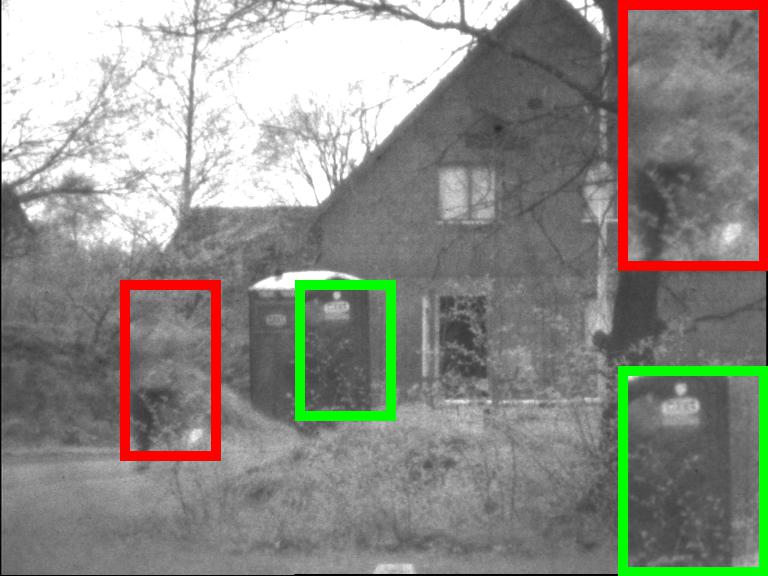}
    \end{subfigure}
    \begin{subfigure}{.095\linewidth}
        \includegraphics[width=\linewidth, height=1.3cm]{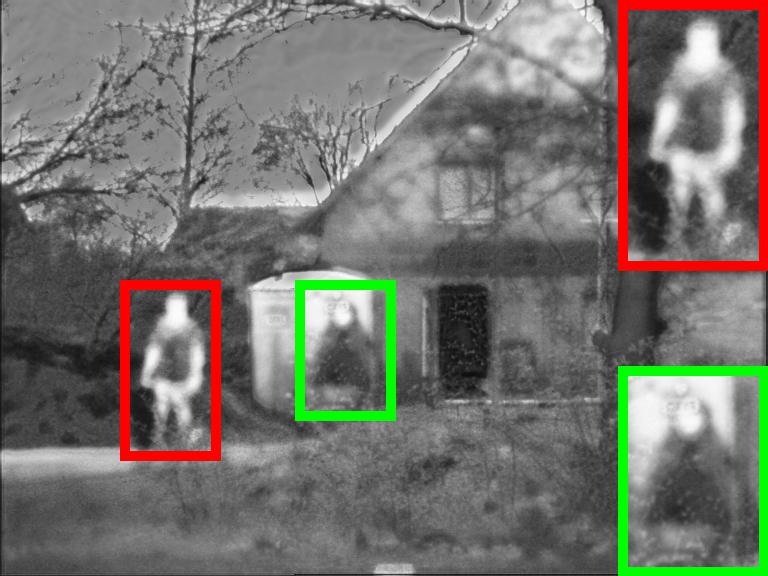}
    \end{subfigure}
    \begin{subfigure}{.095\linewidth}
        \includegraphics[width=\linewidth, height=1.3cm]{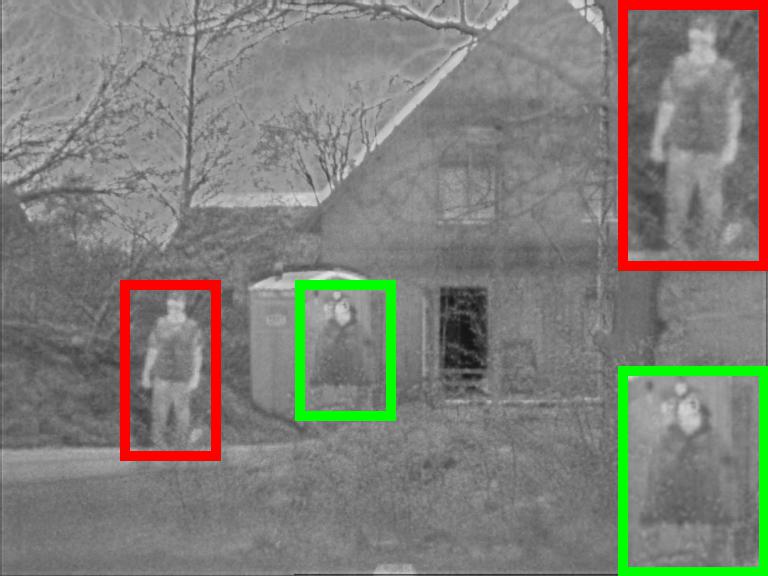}
    \end{subfigure}
    \begin{subfigure}{.095\linewidth}
        \includegraphics[width=\linewidth, height=1.3cm]{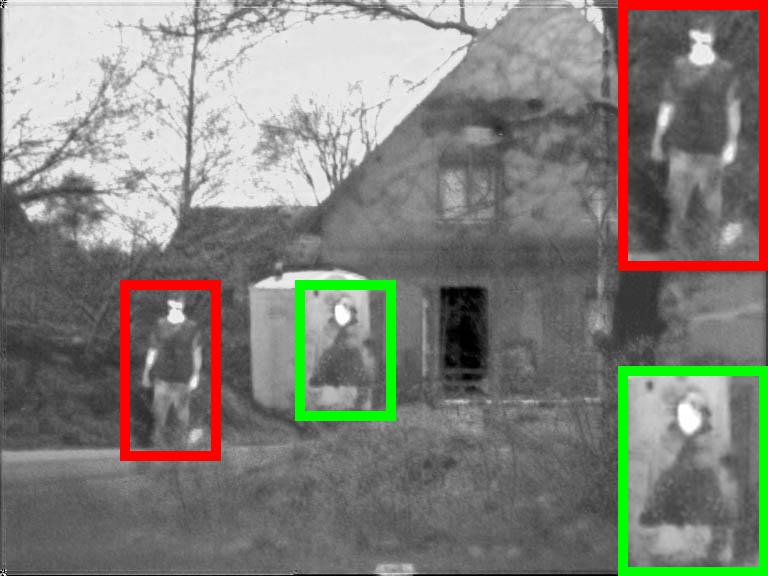}
    \end{subfigure}
    \begin{subfigure}{.095\linewidth}
        \includegraphics[width=\linewidth, height=1.3cm]{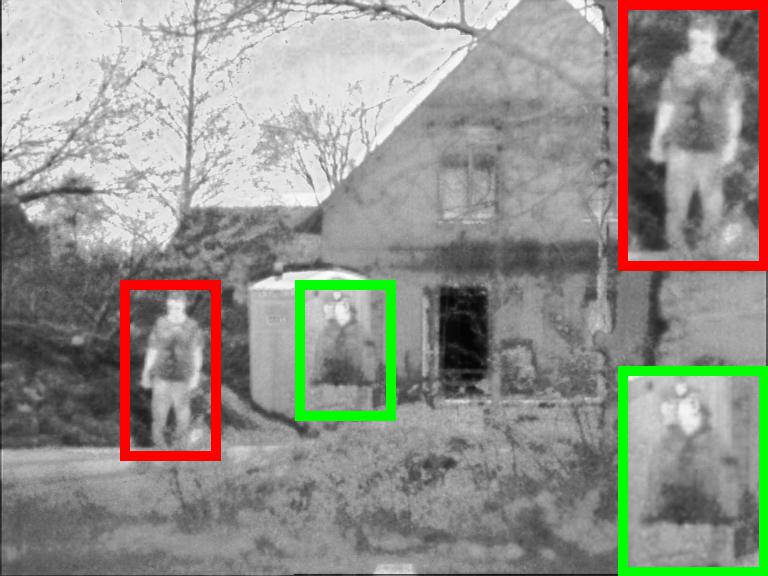}
    \end{subfigure}
    \begin{subfigure}{.095\linewidth}
        \includegraphics[width=\linewidth, height=1.3cm]{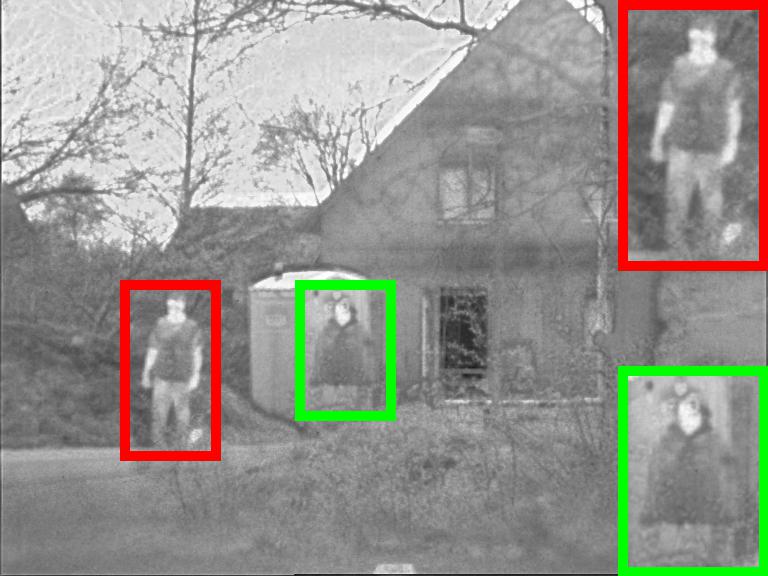}
    \end{subfigure}
    \begin{subfigure}{.095\linewidth}
        \includegraphics[width=\linewidth, height=1.3cm]{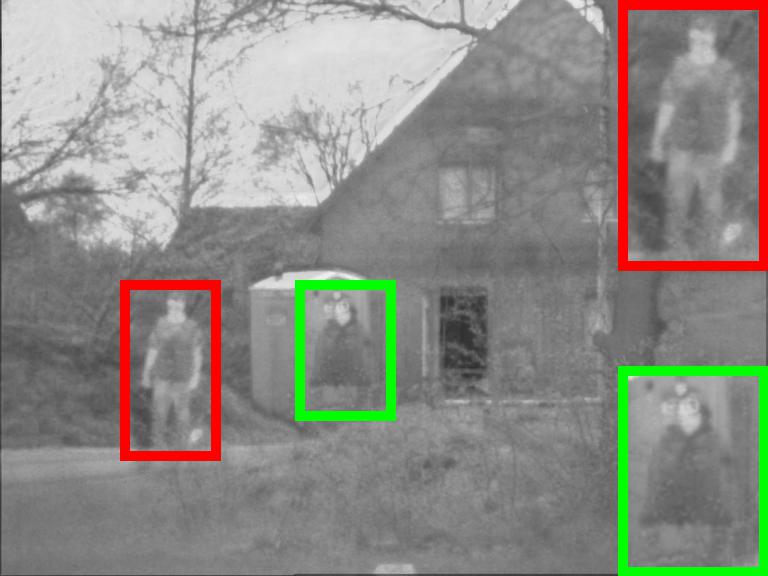}
    \end{subfigure}
    \begin{subfigure}{.095\linewidth}
        \includegraphics[width=\linewidth, height=1.3cm]{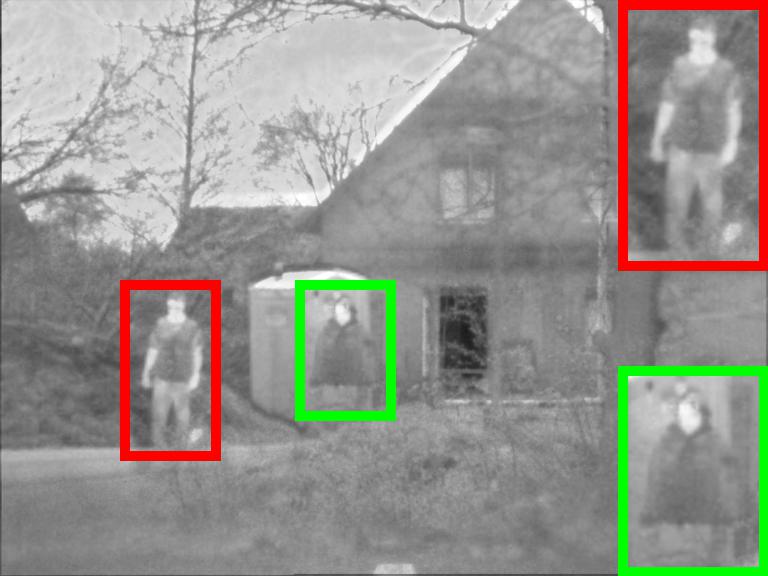}
    \end{subfigure}
    \begin{subfigure}{.095\linewidth}
        \includegraphics[width=\linewidth, height=1.3cm]{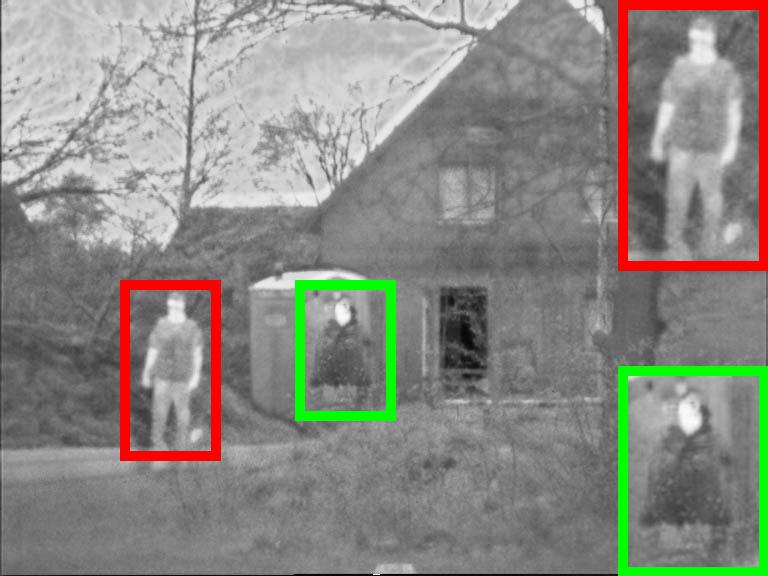}
    \end{subfigure}

    \begin{subfigure}{.095\linewidth}
        \includegraphics[width=\linewidth, height=1.3cm]{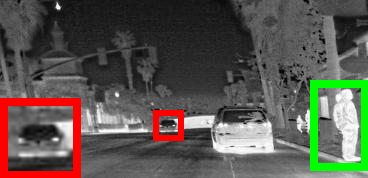}
        \caption*{Ir}
    \end{subfigure}
    \begin{subfigure}{.095\linewidth}
        \includegraphics[width=\linewidth, height=1.3cm]{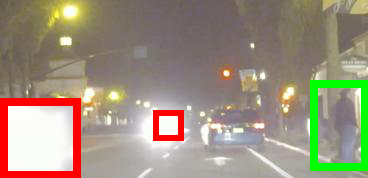}
        \caption*{Vi}
    \end{subfigure}
    \begin{subfigure}{.095\linewidth}
        \includegraphics[width=\linewidth, height=1.3cm]{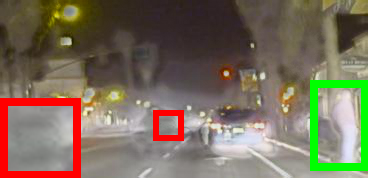}
        \caption*{DDcGAN}
    \end{subfigure}
    \begin{subfigure}{.095\linewidth}
        \includegraphics[width=\linewidth, height=1.3cm]{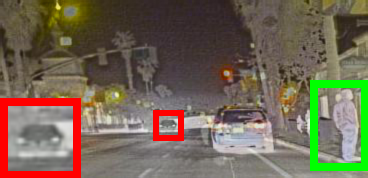}
        \caption*{U2Fusion}
    \end{subfigure}
    \begin{subfigure}{.095\linewidth}
        \includegraphics[width=\linewidth, height=1.3cm]{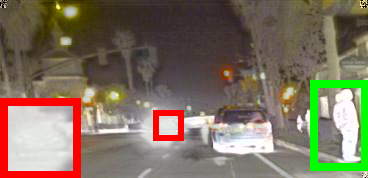}
        \caption*{TarDAL}
    \end{subfigure}
    \begin{subfigure}{.095\linewidth}
        \includegraphics[width=\linewidth, height=1.3cm]{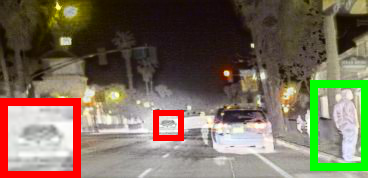}
        \caption*{DDFM}
    \end{subfigure}
    \begin{subfigure}{.095\linewidth}
        \includegraphics[width=\linewidth, height=1.3cm]{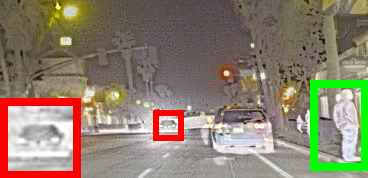}
        \caption*{CDDFuse}
    \end{subfigure}
    \begin{subfigure}{.095\linewidth}
        \includegraphics[width=\linewidth, height=1.3cm]{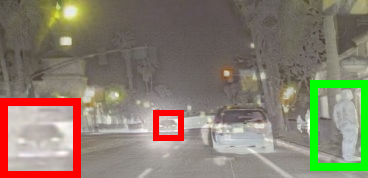}
        \caption*{DeFusion}
    \end{subfigure}
    \begin{subfigure}{.095\linewidth}
        \includegraphics[width=\linewidth, height=1.3cm]{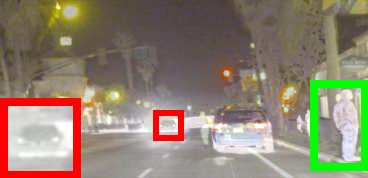}
        \caption*{SegMiF}
    \end{subfigure}
    \begin{subfigure}{.095\linewidth}
        \includegraphics[width=\linewidth, height=1.3cm]{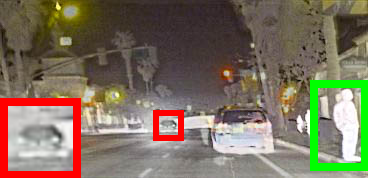}
        \caption*{Ours}
    \end{subfigure}
    \caption{Comparative visual fusion of our proposed method versus state-of-the-art methods on three typical image pairs in \(\text{M}^{3}\text{FD}\), TNO, and RoadScene datasets.}
    \label{exp_fusion_img}
    
\end{figure*}

\subsection{Bilevel Optimizatoin}
In order for our model to delve deeper to requirements of computational perception, we employ a bilevel optimization method as shown in Figure~\ref{pics:net2}. Unlike previous approaches catering for high visual quality, our framework posits that IVIF should yield an image conducive to both human visual assessment and computational perception, specifically object detection. Let the infrared and visible images be represented as gray-scale images of size \( m \times n \), vectorized as \( \mathbf{x} \) and \( \mathbf{y} \) respectively. The fused image is similarly vectorized as \( \mathbf{z} \in \mathbb{R}^{mn \times 1} \). Text prompts are encoded into semantic vectors \( \mathbf{s}_t \) using the CLIP model. Inspired by Stackelberg's theory\cite{BO1,BO2}, we adopt a bilevel optimization framework:
\begin{equation}
\begin{gathered}
\min_{\boldsymbol{\omega}_{\mathrm{d}}} \mathcal{L}^{\mathrm{d}}\left(\Psi\left(\mathbf{z}^* ; \boldsymbol{\omega}_{\mathrm{d}}\right)\right), \\
\text{s.t. } \mathbf{z}^* \in \arg \min_{\mathbf{z}} f(\mathbf{z}, \mathbf{x}, \mathbf{y}, \mathbf{s}_t) + g(\mathbf{z}),
\end{gathered}
\end{equation}

\noindent where \( f(\cdot) \) encapsulates the loss associated with the fusion network, which includes the alignment of the fused image \( \mathbf{z} \) with the source infrared and visible images, as well as the integration of the text prompt semantics. The function \( g(\cdot) \) represents the codebook quantization loss, which quantifies the fidelity of the discrete representation of the fused image in the codebook. For detection, we adopt YOLOv5 as our backbone for the detection network \( \Psi \) with learnable parameters \( \boldsymbol{\omega}_{\mathrm{d}} \), where the detection-specific training loss \(\mathcal{L}^{\mathrm{d}} \) also follows its setting.

In this formulation, the lower-level problem seeks to find an optimal fused image \( \mathbf{z}^* \) by minimizing the fusion and codebook losses. The upper-level problem then optimizes the parameters of the detection network to minimize the detection loss, given the optimal fused image from the lower level. This ensures that the fused image is conducive to both visual quality and detection efficacy, thereby serving the dual purposes of human and computer vision.

The structure loss plays a pivotal role in leveraging the information encapsulated in the intermediate fusion results to enhance the training process across epochs. The mathematical expression of the structure loss is given by:
\begin{equation}
\mathcal{L}^{str} = \mathcal{L}^{SSIM}\left(\mathbf{z}, \mathbf{z}^{\prime}\right) + \mathcal{L}^{pixel}\left(\mathbf{z}, \mathbf{z}^{\prime}\right) + \mathcal{L}^{ grad}\left(\mathbf{z}, \mathbf{z}^{\prime}\right),
\end{equation}
\noindent with \(\mathcal{L}^{\text{SSIM}}\), \(\mathcal{L}^{\text{pixel}}\), and \(\mathcal{L}^{\text{grad}}\) representing the structural similarity loss, pixel intensity loss, and gradient loss, respectively. The incorporation of the structure loss in our framework ensures a progressive refinement of the fusion quality, thereby establishing an evolutionary training paradigm that systematically exploits the accumulated knowledge of the network across epochs.

The content consistency loss is designed to preserve the essential attributes of the source imagery in the fused output. Specifically, the content consistency loss is computed based on the saliency degree weight. Supposing that the saliency value of \( \mathbf{x} \) at the \(k^{th}\) pixel can be obtained by $\boldsymbol{S}_{\mathbf{x}(k)}=\sum_{i=0}^{255} \boldsymbol{H}_{\mathbf{x}}(i)|\mathbf{x}(k)-i|$, where \( \mathbf{x}(k) \)is the value of the \(k^{th}\) pixel and \(\boldsymbol{H}_{\mathbf{x}}(i)\) is the histogram of pixel value \(i\), the formulation of the content consistency loss is as follows:
\begin{equation}
\mathcal{L}^{cc} = \sum_{i=1}^2 \bigg(\mathcal{L}^{{t \in\{SSIM, pixel, grad\}}}(\mathbf{z}, \omega_i I_i)\bigg),
\end{equation}

\noindent where \(\omega_1 = \frac{S_x(k)}{|S_x(k) - S_y(k)|}\), and \(\omega_2 = 1 - \omega_1\). Here \(I_1\) and \(I_2\) represent the input infrared and visible images, respectively.

The feasibility constraint of the fusion network is the combination of the aforementioned two main parts:
\begin{equation}
\mathcal{L}^{f} = \alpha_1 \mathcal{L}^{str}\left(\mathbf{z}, \mathbf{z}^{\prime}\right)+\alpha_2 \mathcal{L}^{cc}(\mathbf{x}, \mathbf{y}, \mathbf{z}),
\end{equation}

\noindent where the \(\alpha_1\) is the average SSIM between \(\mathbf{z}^{\prime}\) and \(\mathbf{x}\), \(\mathbf{y}\). \(\alpha_2\) is the average SSIM between \(\mathbf{z}\) and \(\mathbf{x}\), \(\mathbf{y}\).

We employ a quantization loss function, which quantifies the discrepancy between the original fused feature vectors and their quantized counterparts as follows:
\begin{equation}
g(\mathbf{z}) = \frac{1}{N} \sum_{i=1}^{N} \left\| \mathbf{z}_i - q(\mathbf{z}_i) \right\|^2,
\end{equation}

\noindent where \( \mathbf{z}_i \) is the fused feature vector, \( q(\mathbf{z}_i) \) is the quantized vector retrieved from the codebook, and \( N \) is the batch size. The minimization of \( g \) ensures integrity of the feature space within the constraints of a finite codebook.


\begin{figure*}[ht!]
\setlength{\belowcaptionskip}{-10pt} 

\setlength{\intextsep}{5pt} 
    \centering
    \begin{subfigure}{.12\linewidth}
        \includegraphics[width=\linewidth]{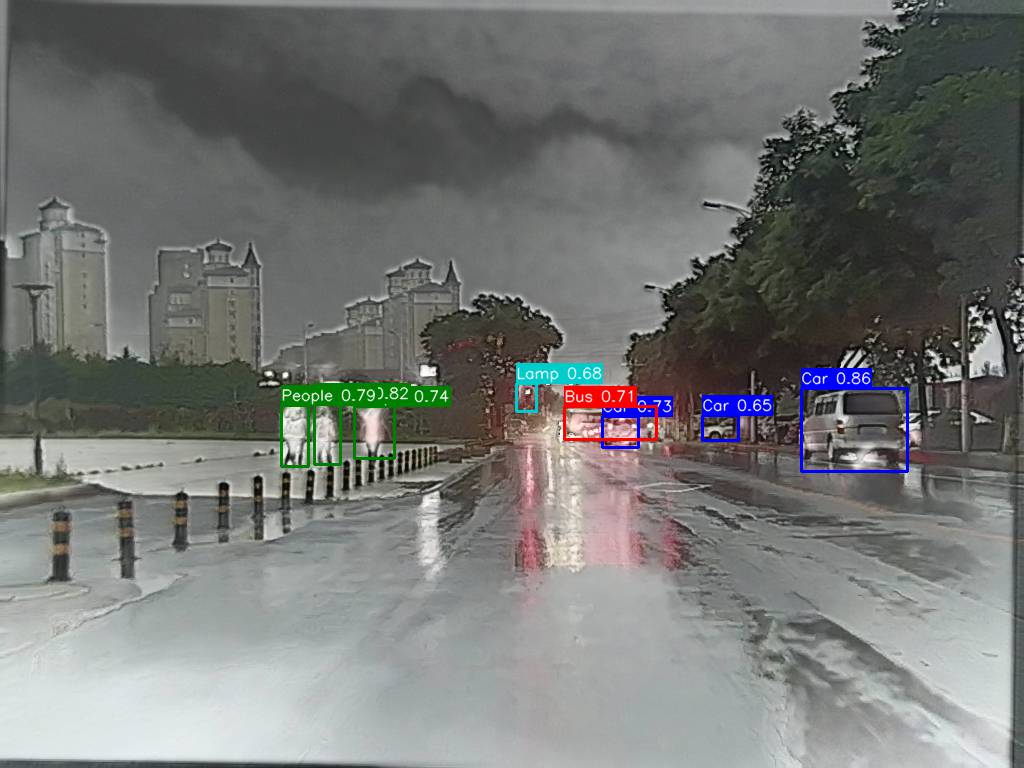}
    \end{subfigure}
    \begin{subfigure}{.12\linewidth}
        \includegraphics[width=\linewidth]{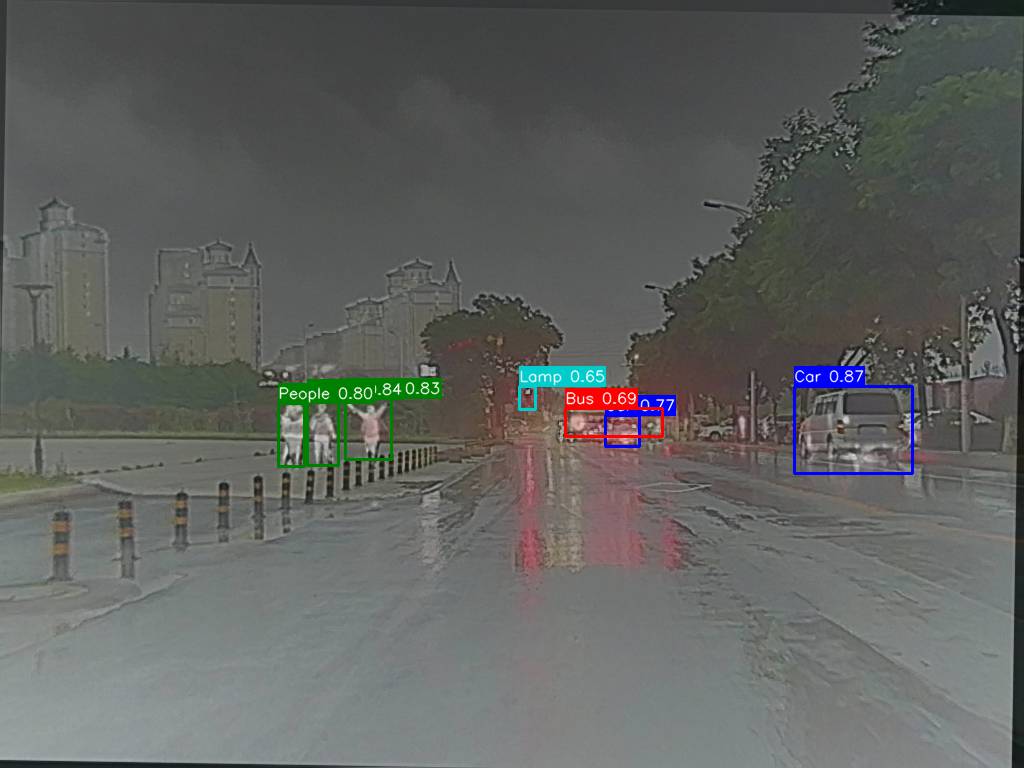}
    \end{subfigure}
    \begin{subfigure}{.12\linewidth}
        \includegraphics[width=\linewidth]{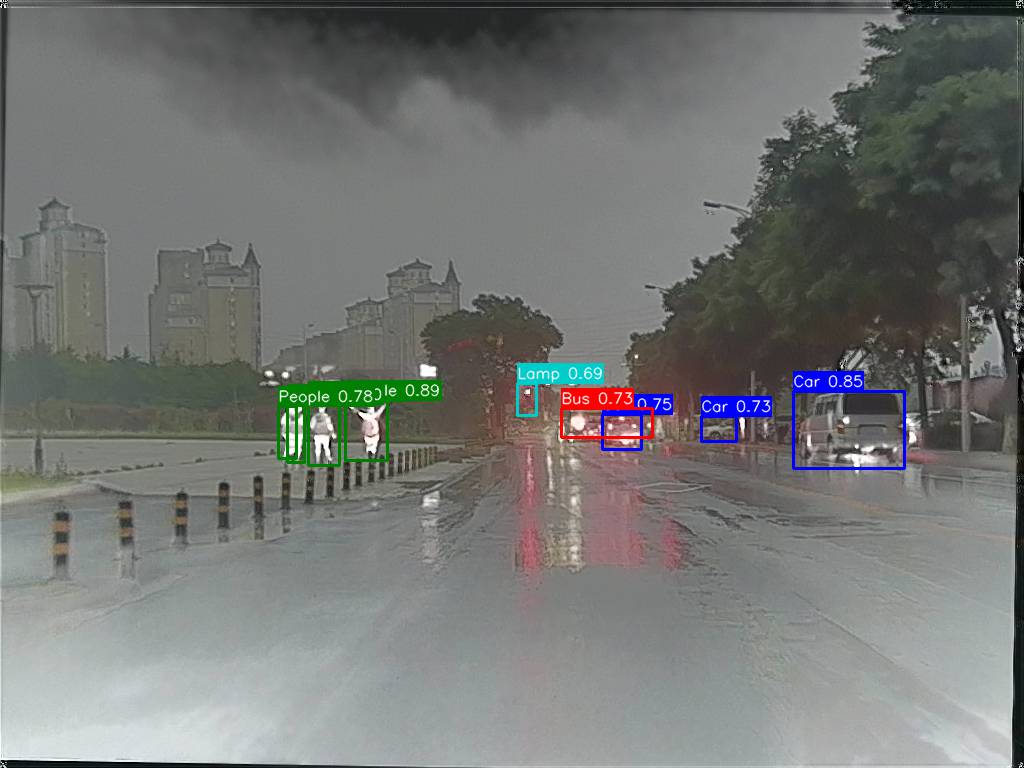}
    \end{subfigure}
    \begin{subfigure}{.12\linewidth}
        \includegraphics[width=\linewidth]{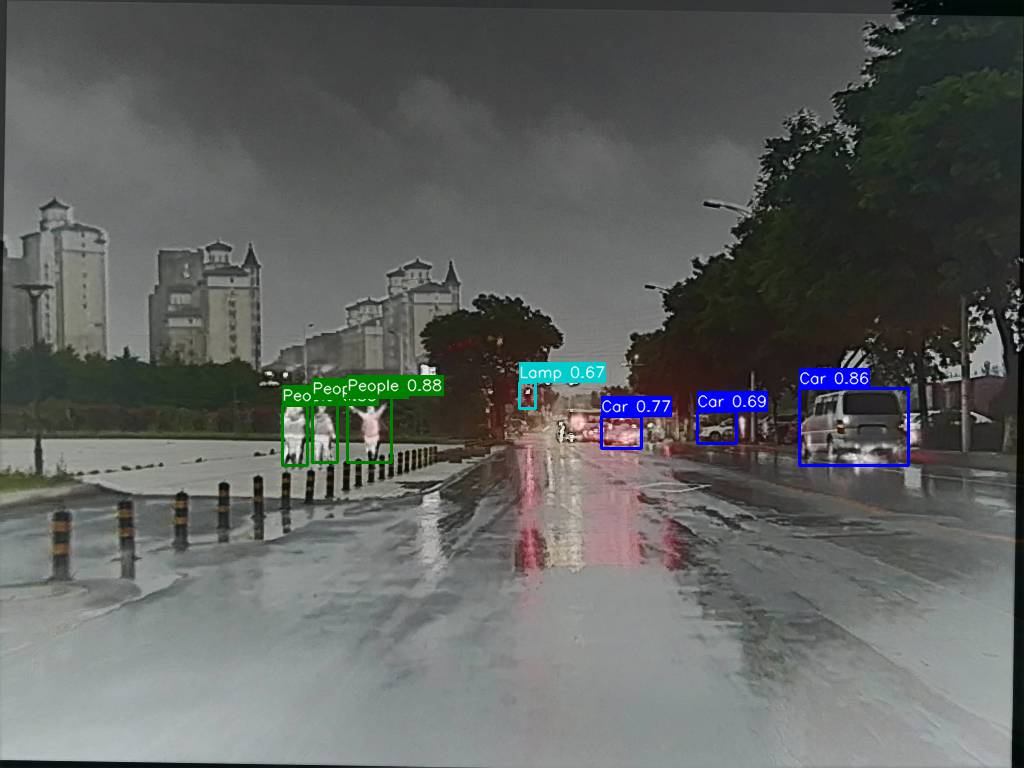}
    \end{subfigure}
    \begin{subfigure}{.12\linewidth}
        \includegraphics[width=\linewidth]{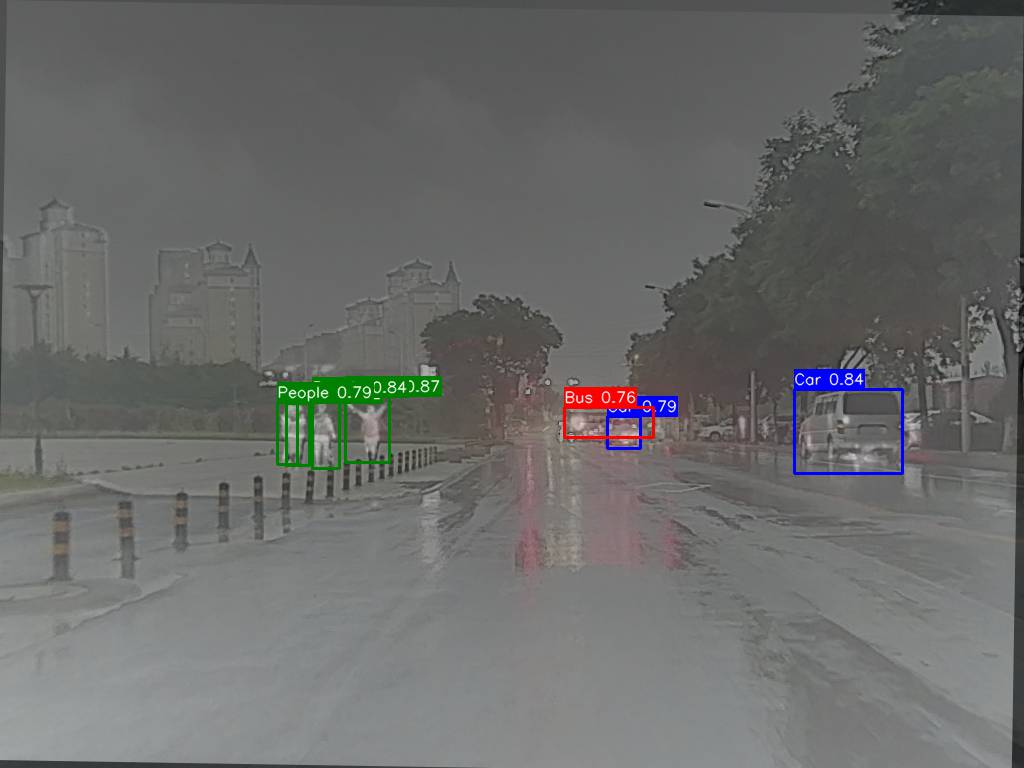}
    \end{subfigure}
    \begin{subfigure}{.12\linewidth}
        \includegraphics[width=\linewidth]{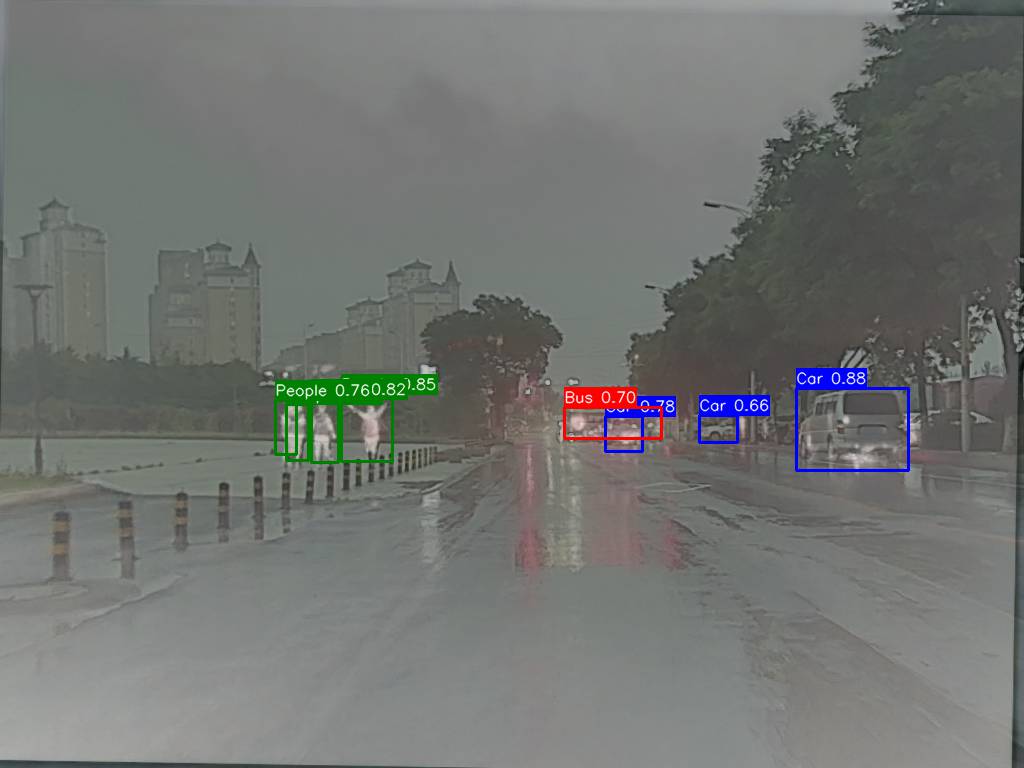}
    \end{subfigure}
    \begin{subfigure}{.12\linewidth}
        \includegraphics[width=\linewidth]{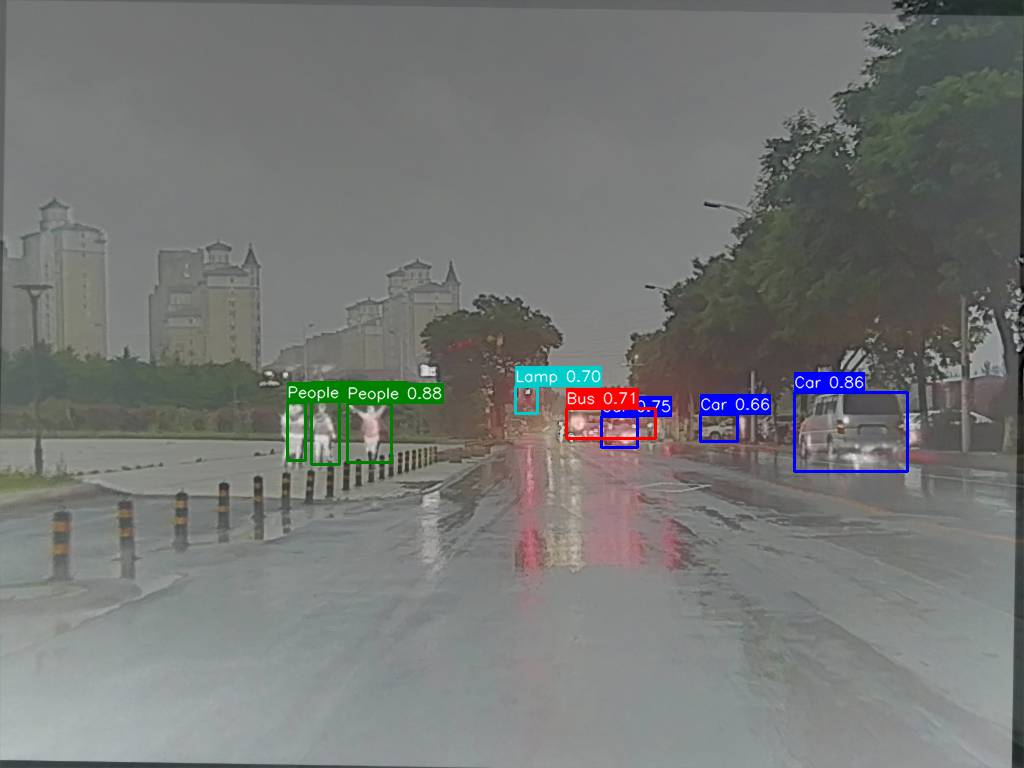}
    \end{subfigure}
    \begin{subfigure}{.12\linewidth}
        \includegraphics[width=\linewidth]{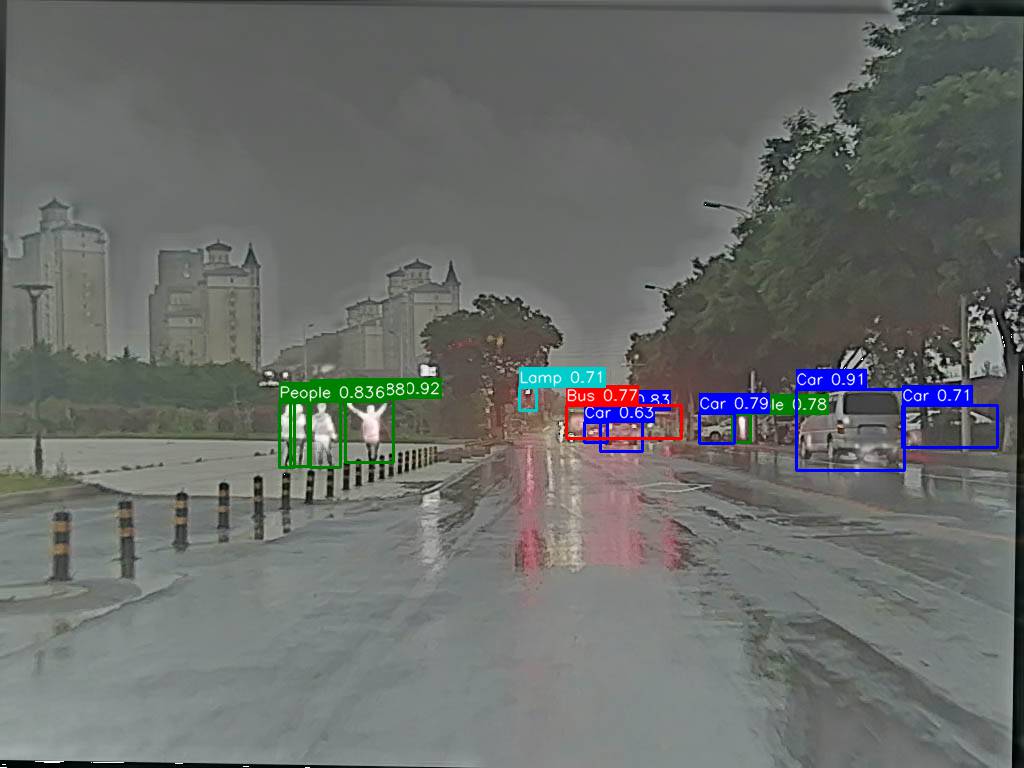}
    \end{subfigure}

    \begin{subfigure}{.12\linewidth}
        \includegraphics[width=\linewidth]{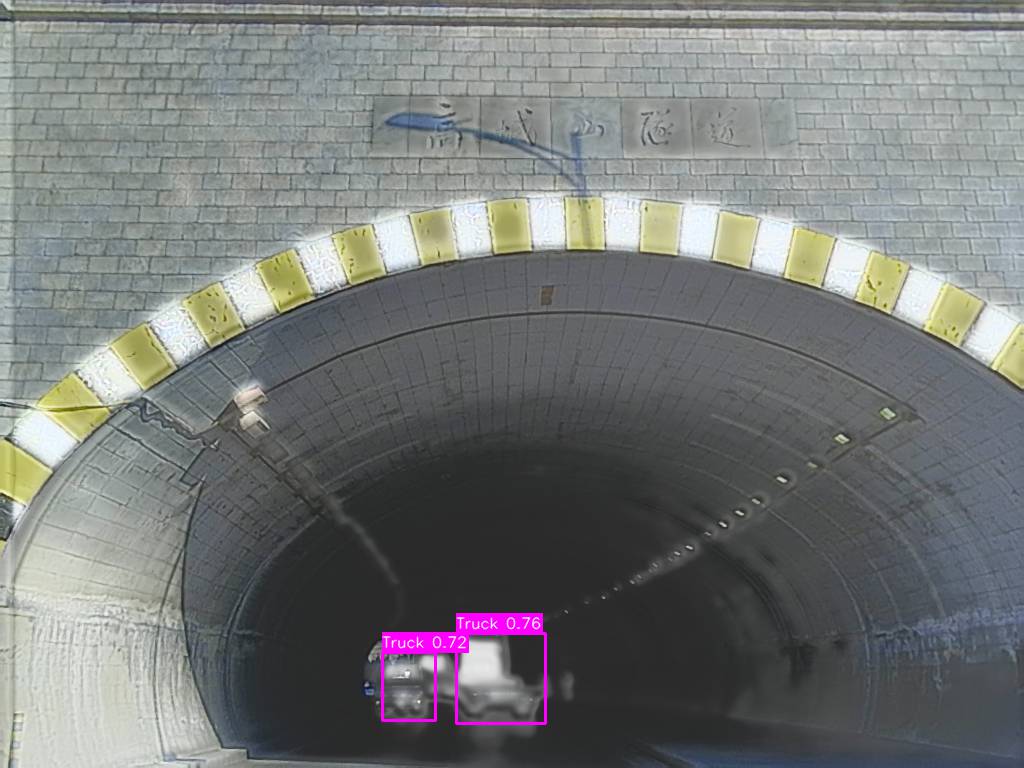}
    \end{subfigure}
    \begin{subfigure}{.12\linewidth}
        \includegraphics[width=\linewidth]{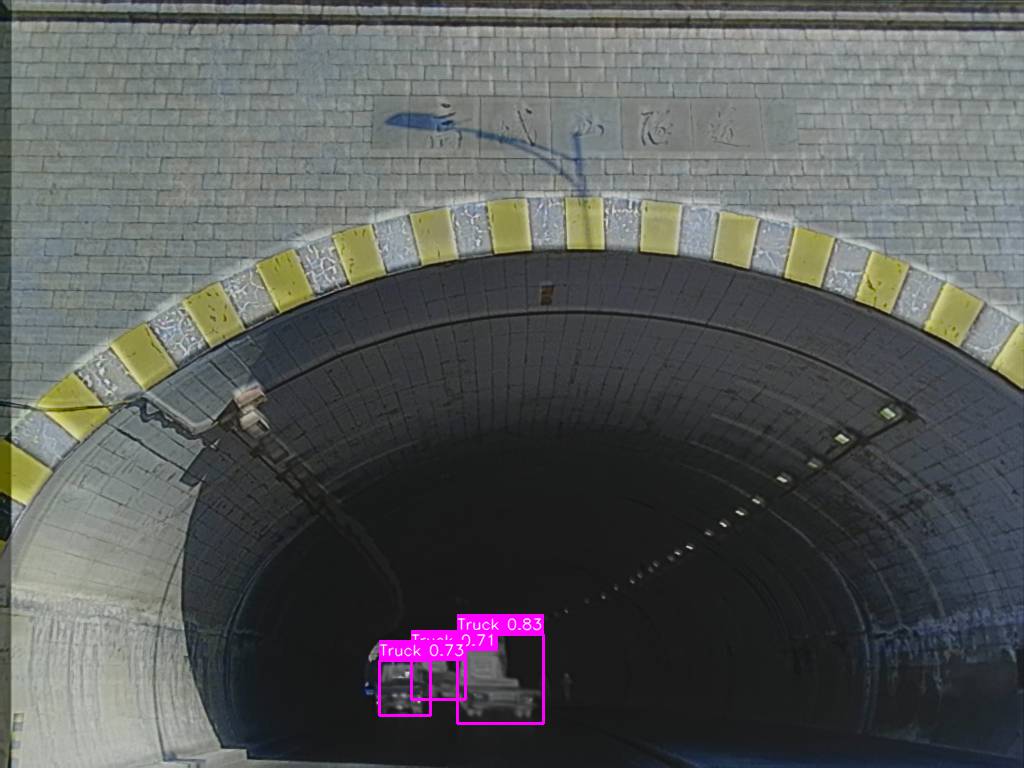}
    \end{subfigure}
    \begin{subfigure}{.12\linewidth}
        \includegraphics[width=\linewidth]{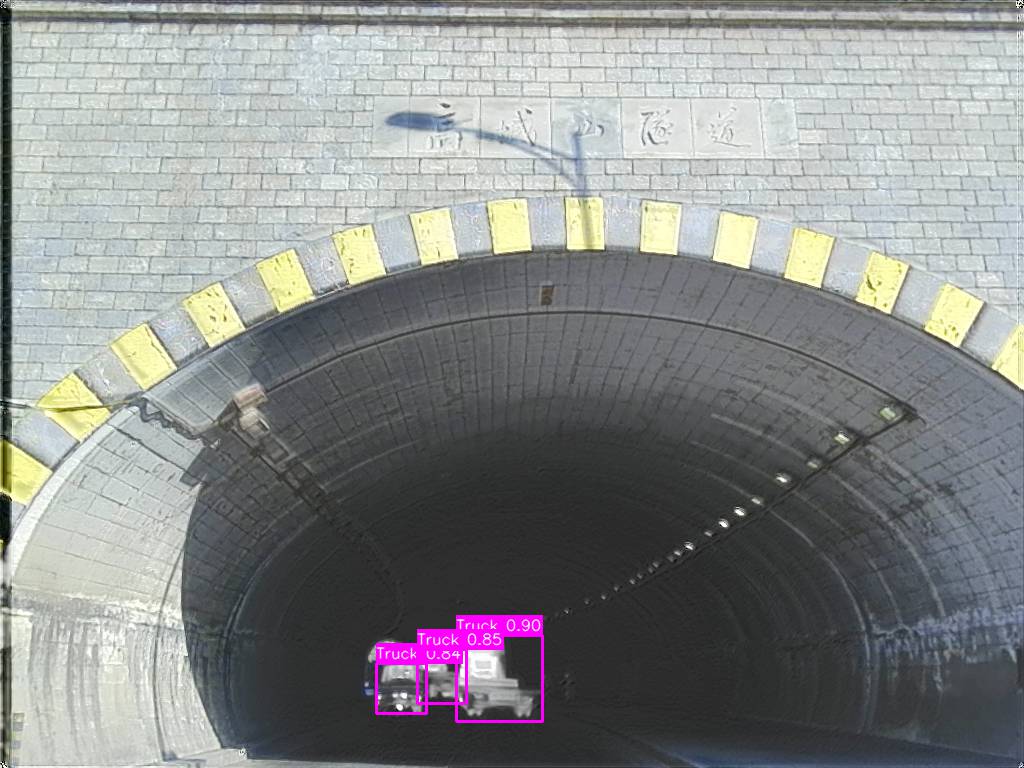}
    \end{subfigure}
    \begin{subfigure}{.12\linewidth}
        \includegraphics[width=\linewidth]{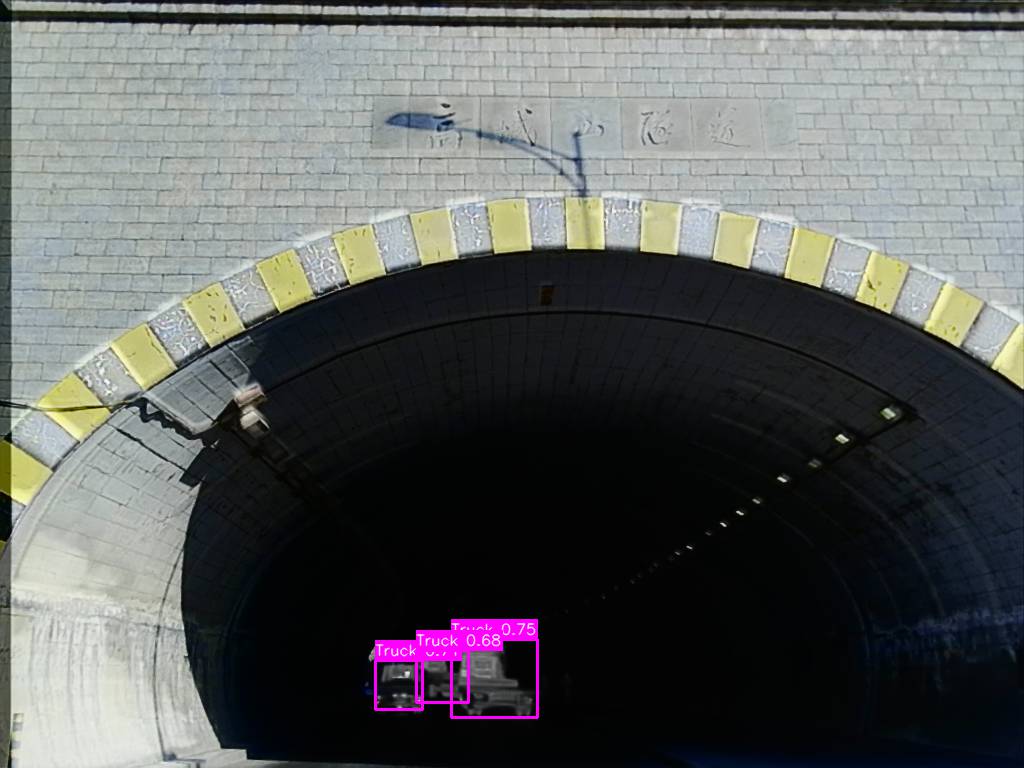}
    \end{subfigure}
    \begin{subfigure}{.12\linewidth}
        \includegraphics[width=\linewidth]{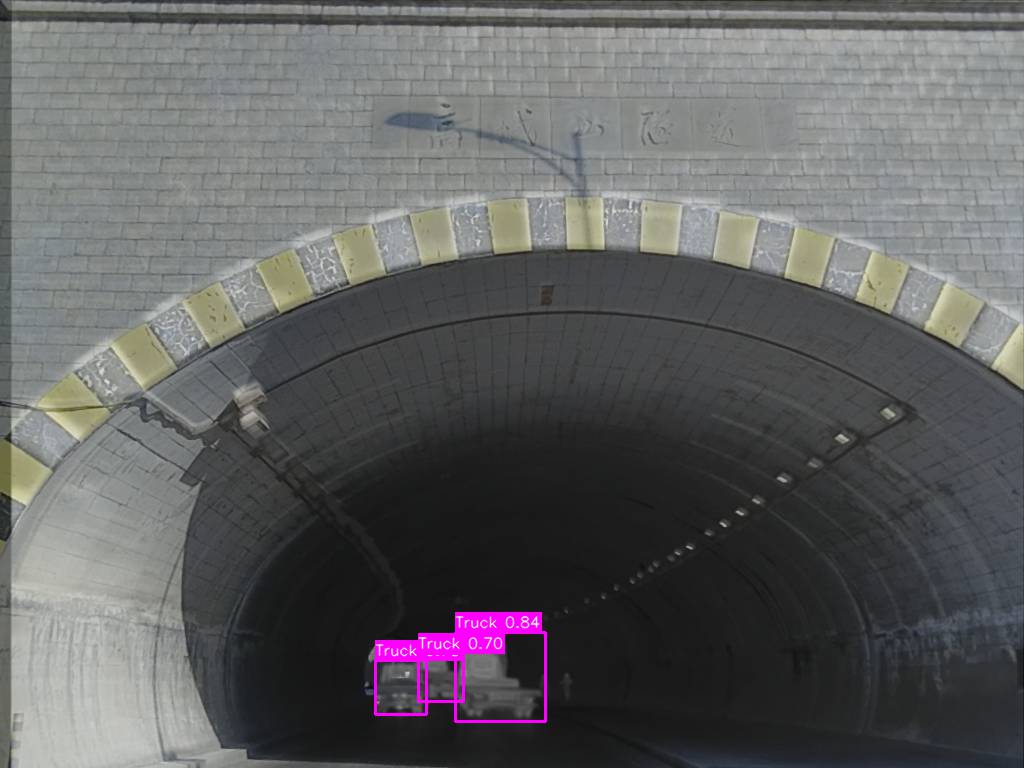}
    \end{subfigure}
    \begin{subfigure}{.12\linewidth}
        \includegraphics[width=\linewidth]{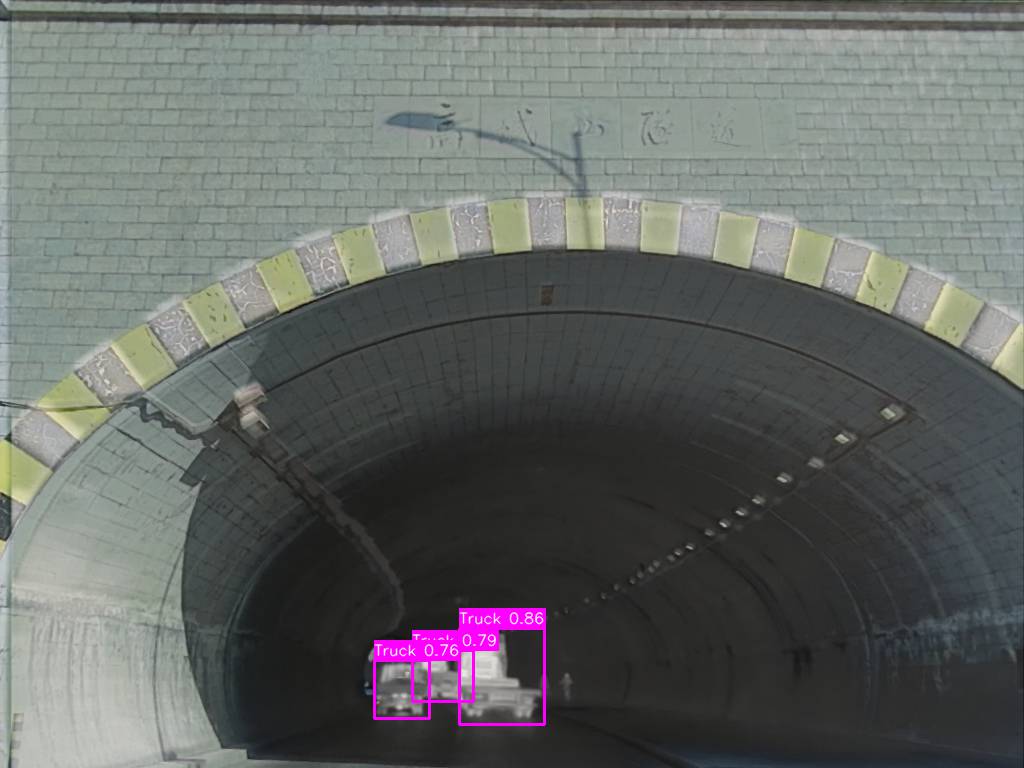}
    \end{subfigure}
    \begin{subfigure}{.12\linewidth}
        \includegraphics[width=\linewidth]{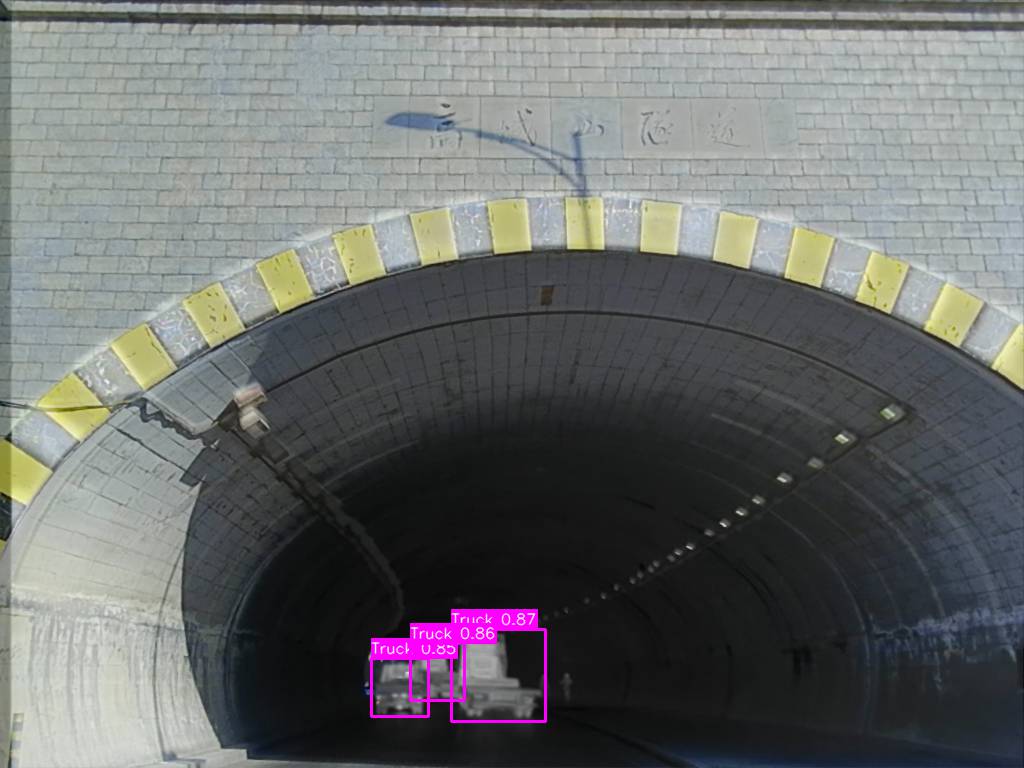}
    \end{subfigure}
    \begin{subfigure}{.12\linewidth}
        \includegraphics[width=\linewidth]{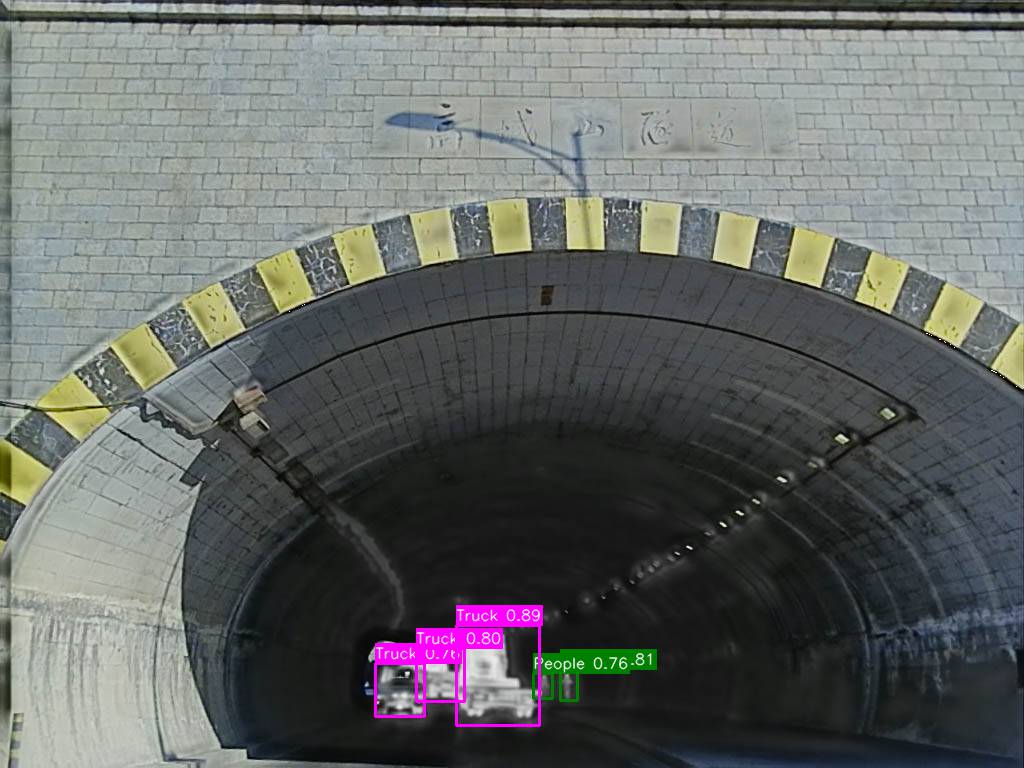}
    \end{subfigure}

    \begin{subfigure}{.12\linewidth}
        \includegraphics[width=\linewidth]{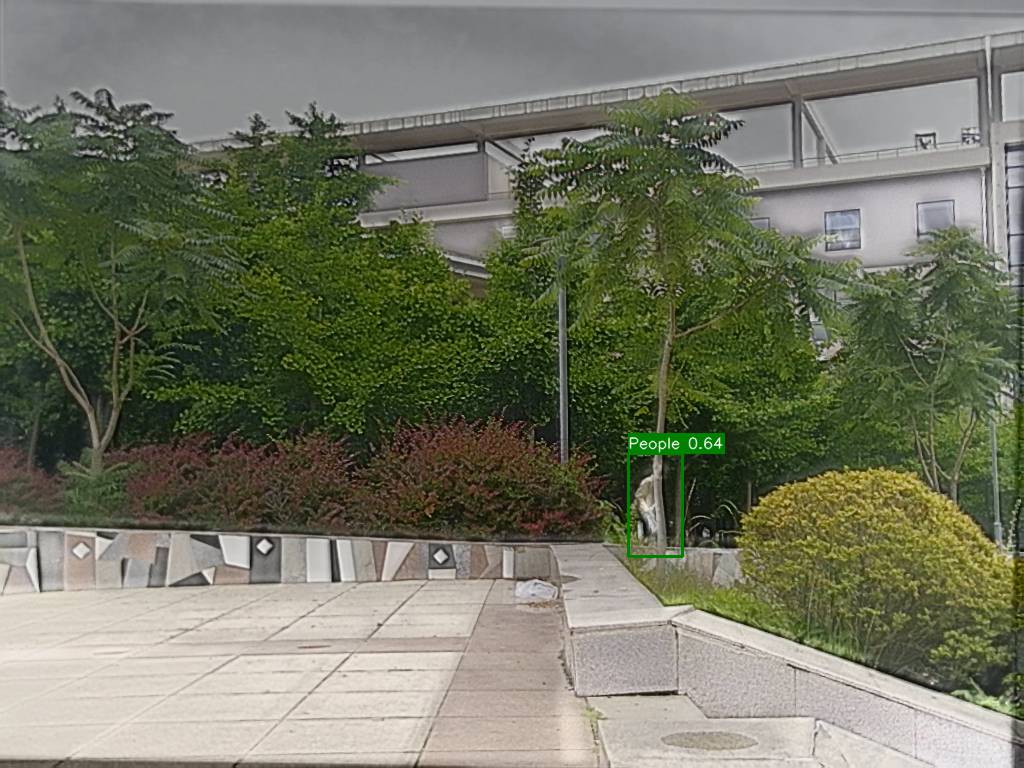}
    \end{subfigure}
    \begin{subfigure}{.12\linewidth}
        \includegraphics[width=\linewidth]{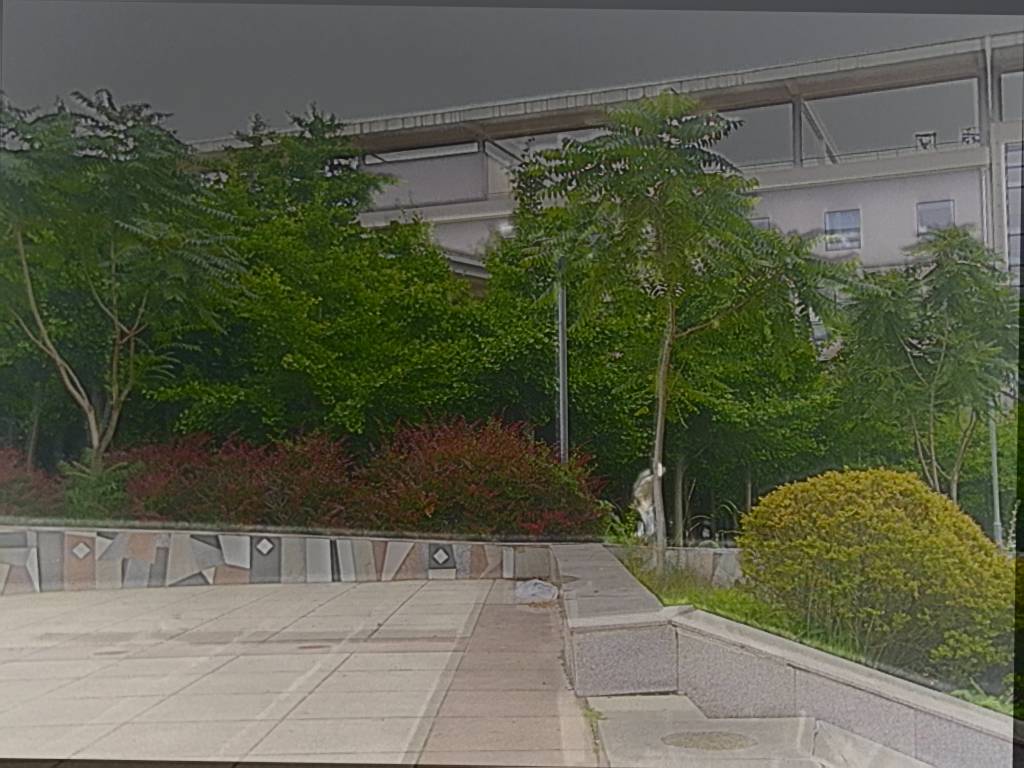}
    \end{subfigure}
    \begin{subfigure}{.12\linewidth}
        \includegraphics[width=\linewidth]{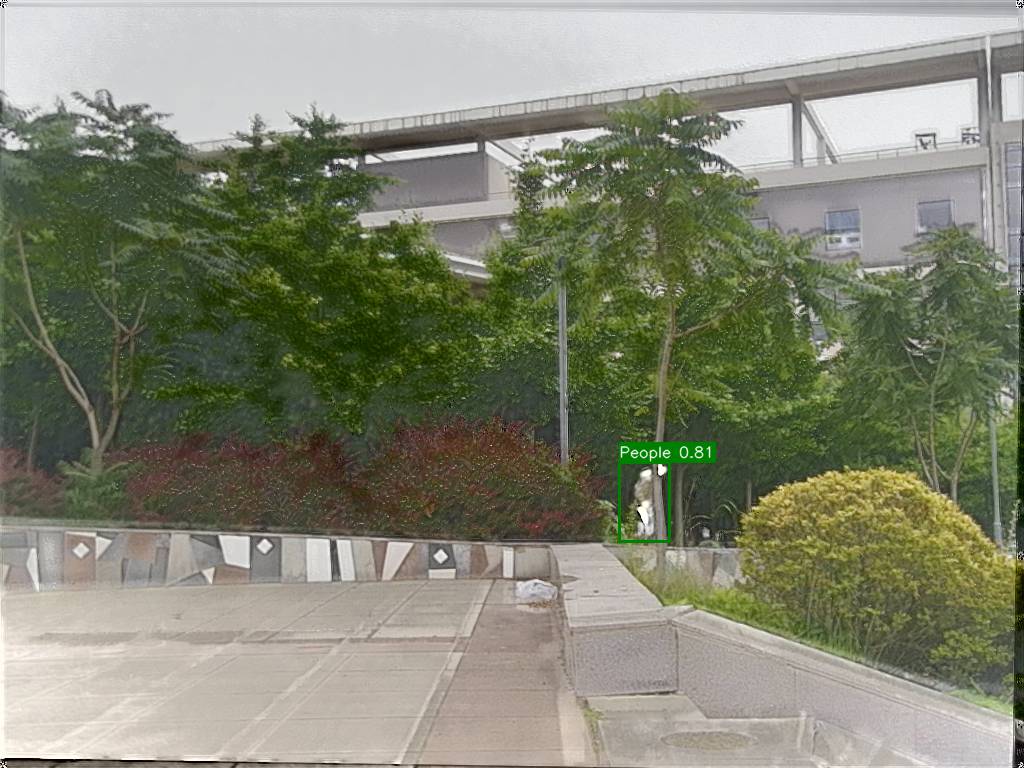}
    \end{subfigure}
    \begin{subfigure}{.12\linewidth}
        \includegraphics[width=\linewidth]{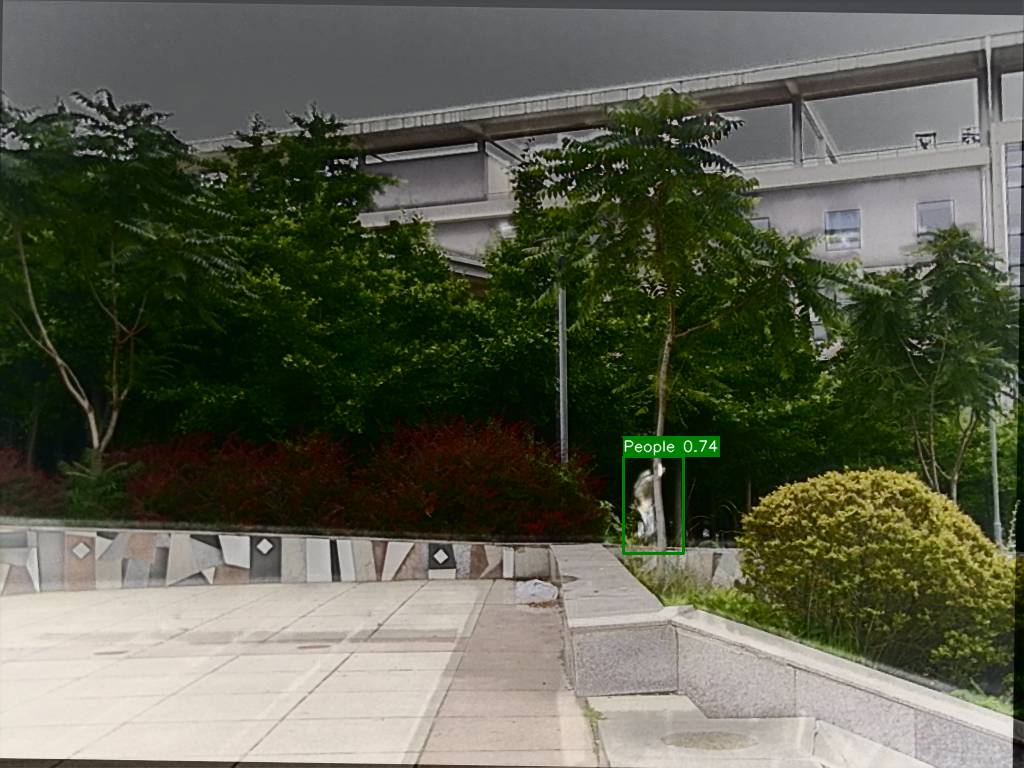}
    \end{subfigure}
    \begin{subfigure}{.12\linewidth}
        \includegraphics[width=\linewidth]{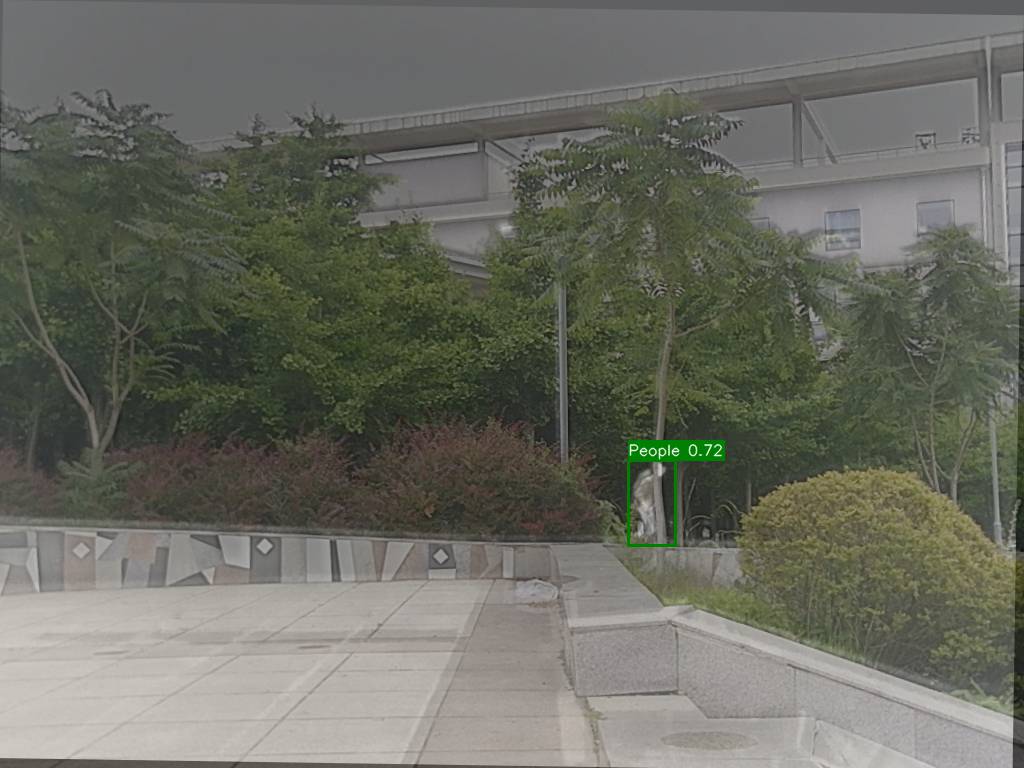}
    \end{subfigure}
    \begin{subfigure}{.12\linewidth}
        \includegraphics[width=\linewidth]{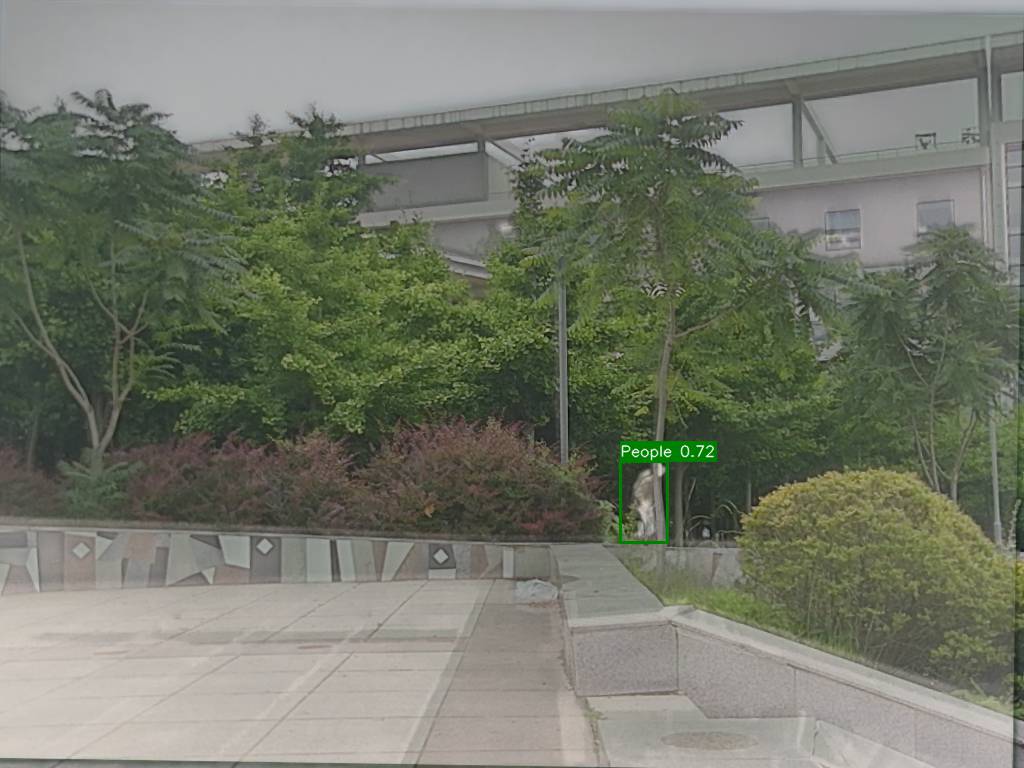}
    \end{subfigure}
    \begin{subfigure}{.12\linewidth}
        \includegraphics[width=\linewidth]{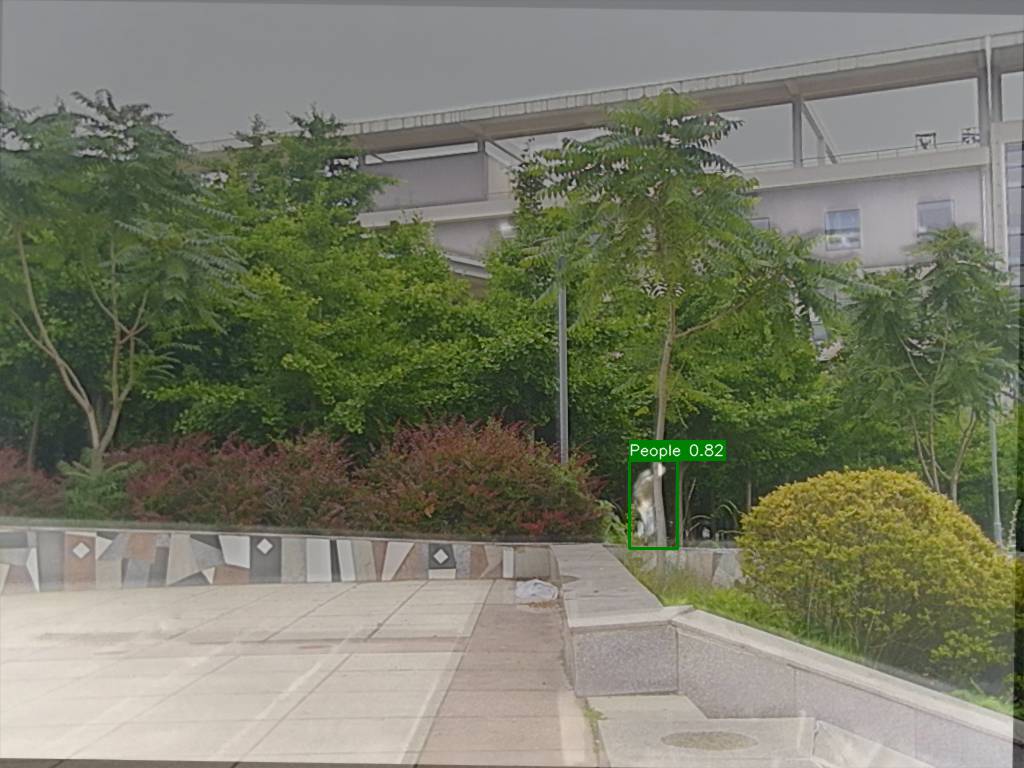}
    \end{subfigure}
    \begin{subfigure}{.12\linewidth}
        \includegraphics[width=\linewidth]{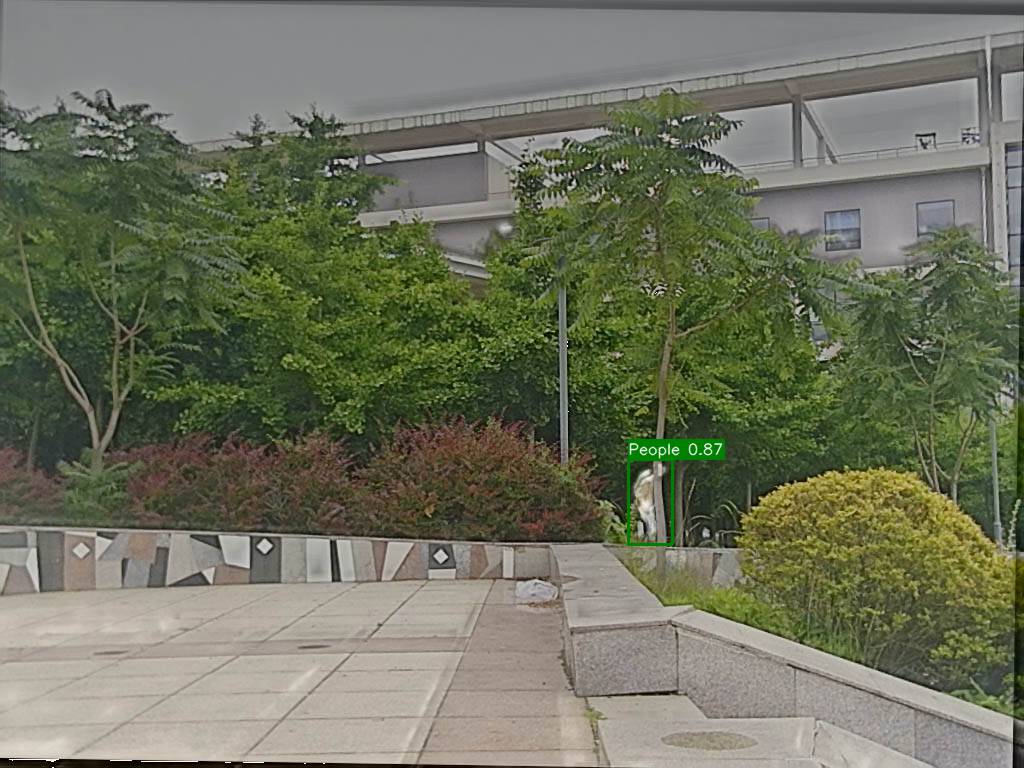}
    \end{subfigure}

    \begin{subfigure}{.12\linewidth}
        \includegraphics[width=\linewidth]{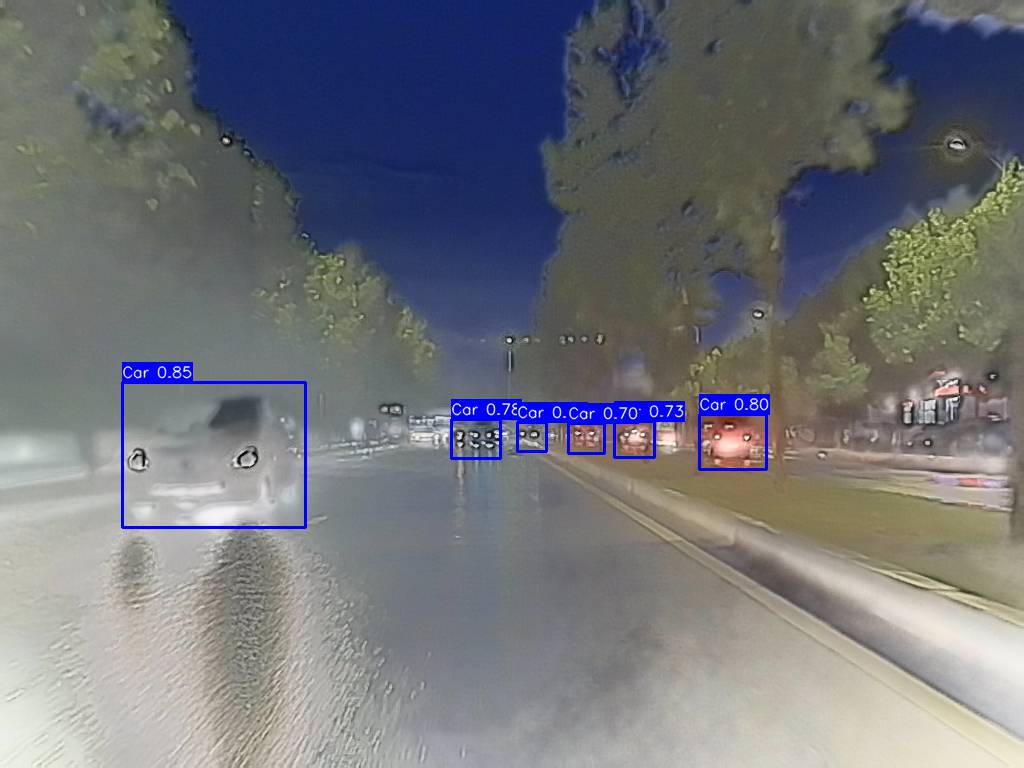}
        \caption*{DDcGAN}
    \end{subfigure}
    \begin{subfigure}{.12\linewidth}
        \includegraphics[width=\linewidth]{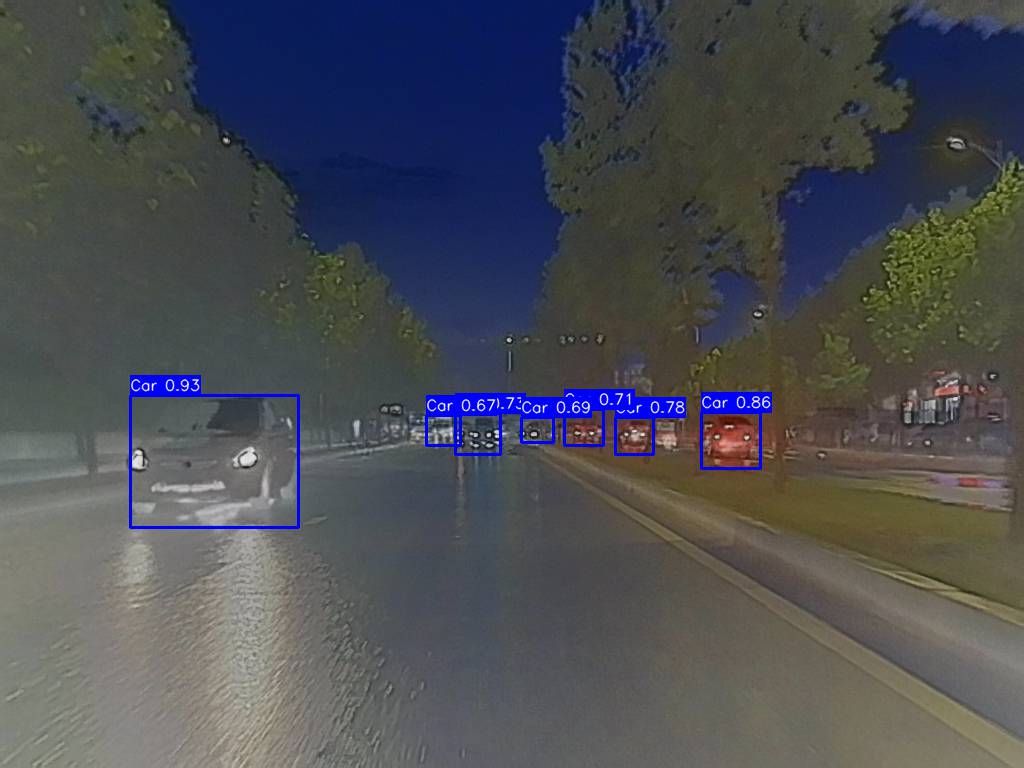}
        \caption*{U2Fusion}
    \end{subfigure}
    \begin{subfigure}{.12\linewidth}
        \includegraphics[width=\linewidth]{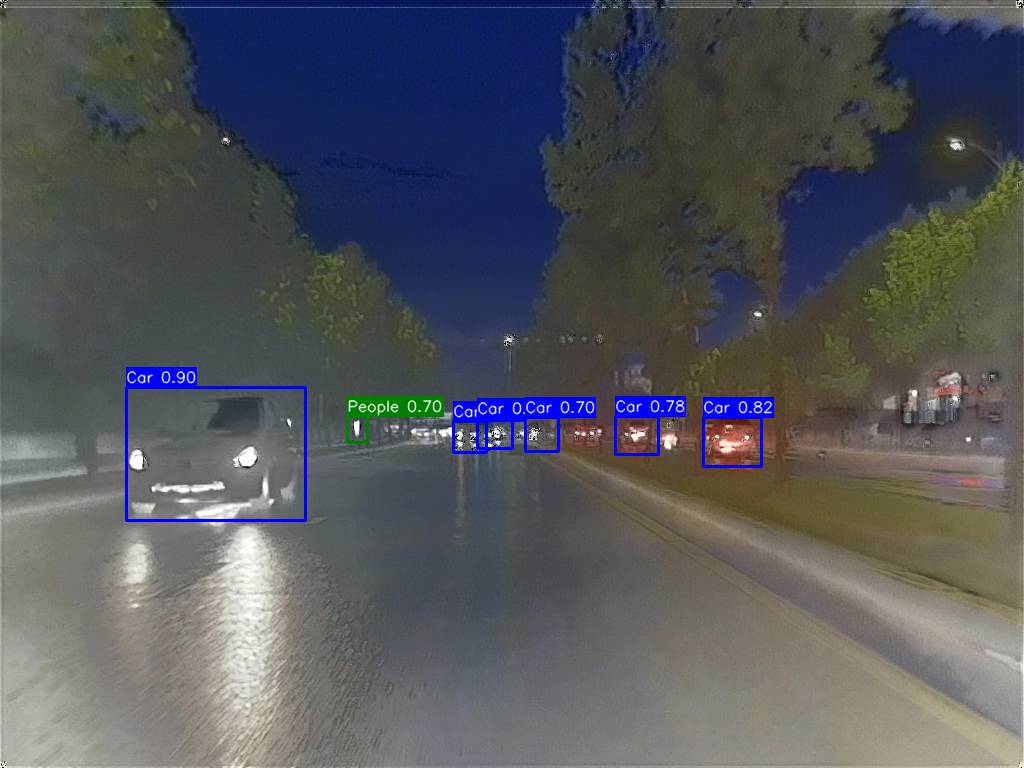}
        \caption*{TarDAL}
    \end{subfigure}
    \begin{subfigure}{.12\linewidth}
        \includegraphics[width=\linewidth]{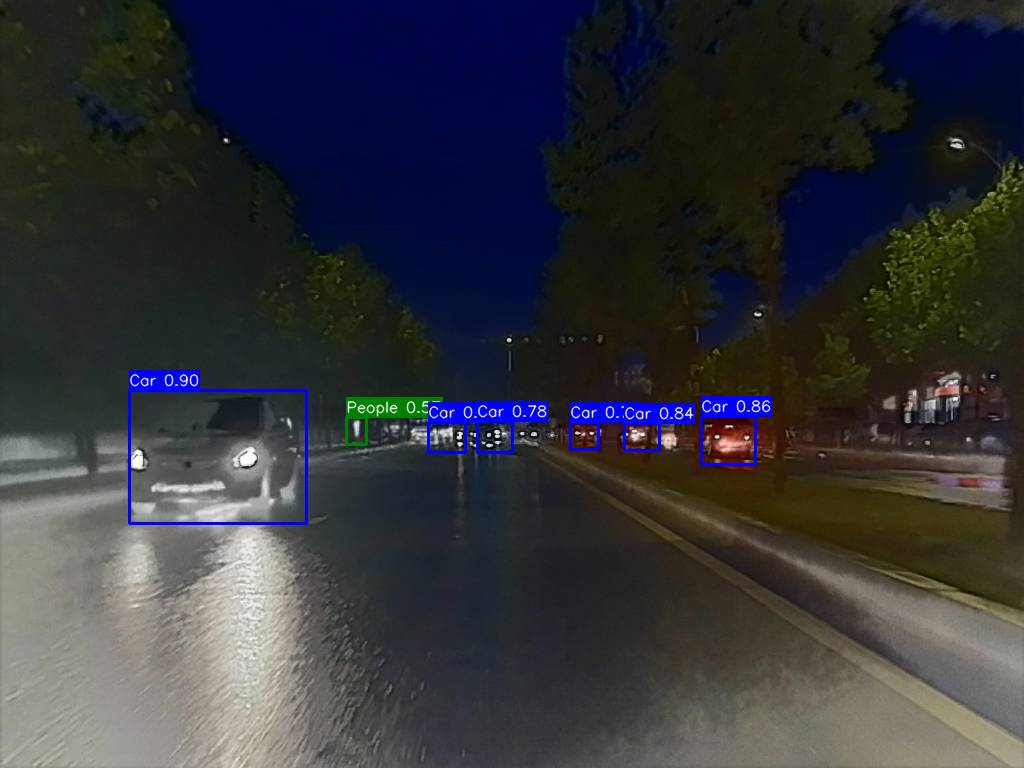}
        \caption*{DDFM}
    \end{subfigure}
    \begin{subfigure}{.12\linewidth}
        \includegraphics[width=\linewidth]{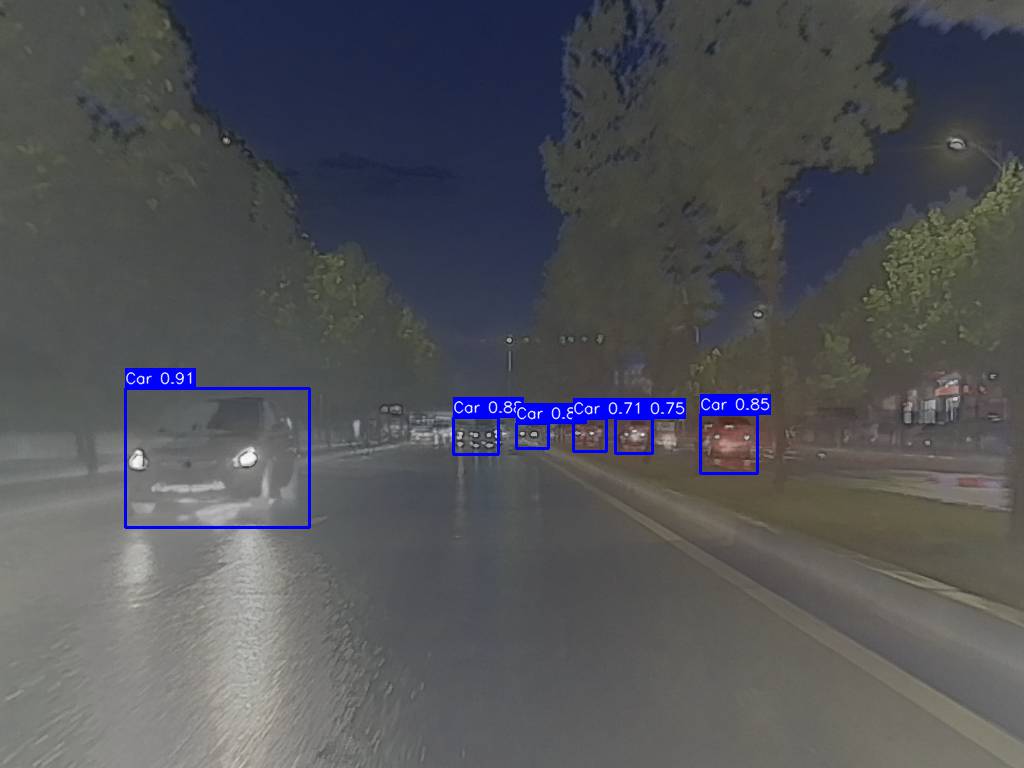}
        \caption*{CDDFuse}
    \end{subfigure}
    \begin{subfigure}{.12\linewidth}
        \includegraphics[width=\linewidth]{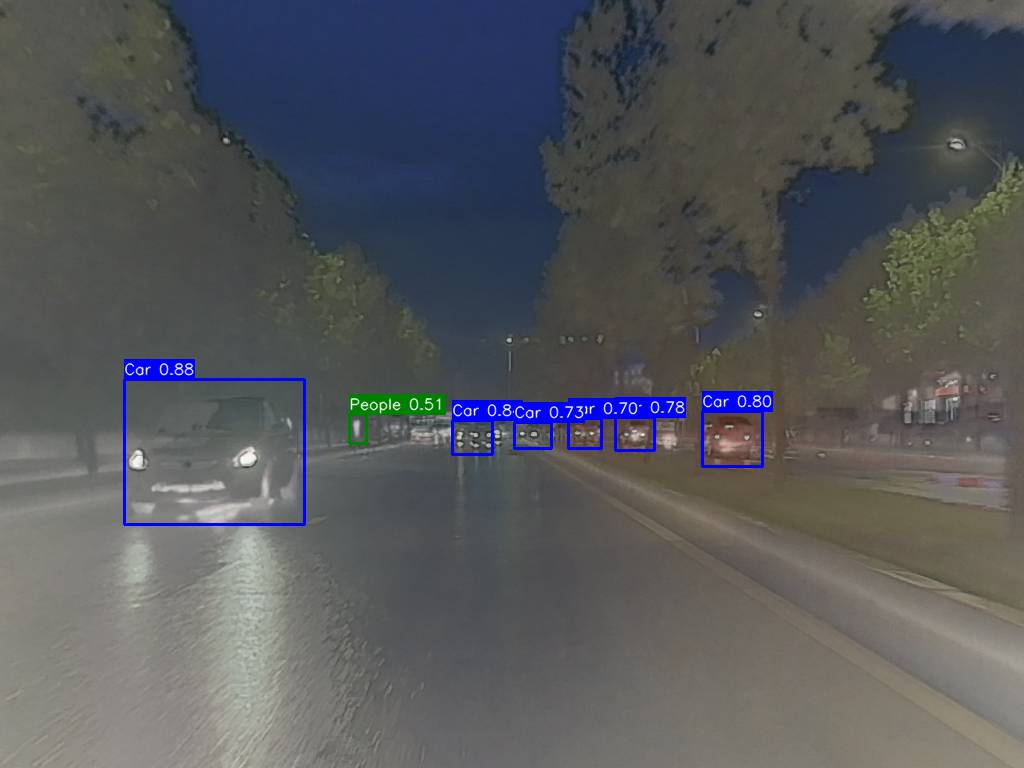}
        \caption*{DeFusion}
    \end{subfigure}
    \begin{subfigure}{.12\linewidth}
        \includegraphics[width=\linewidth]{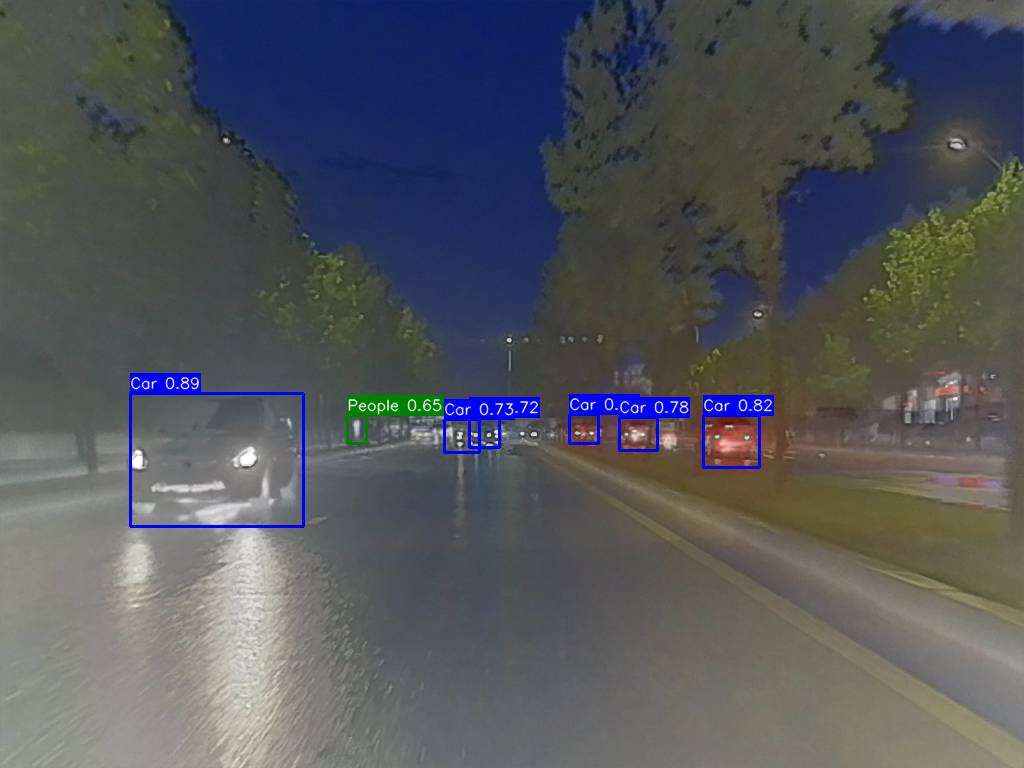}
        \caption*{SegMiF}
    \end{subfigure}
    \begin{subfigure}{.12\linewidth}
        \includegraphics[width=\linewidth]{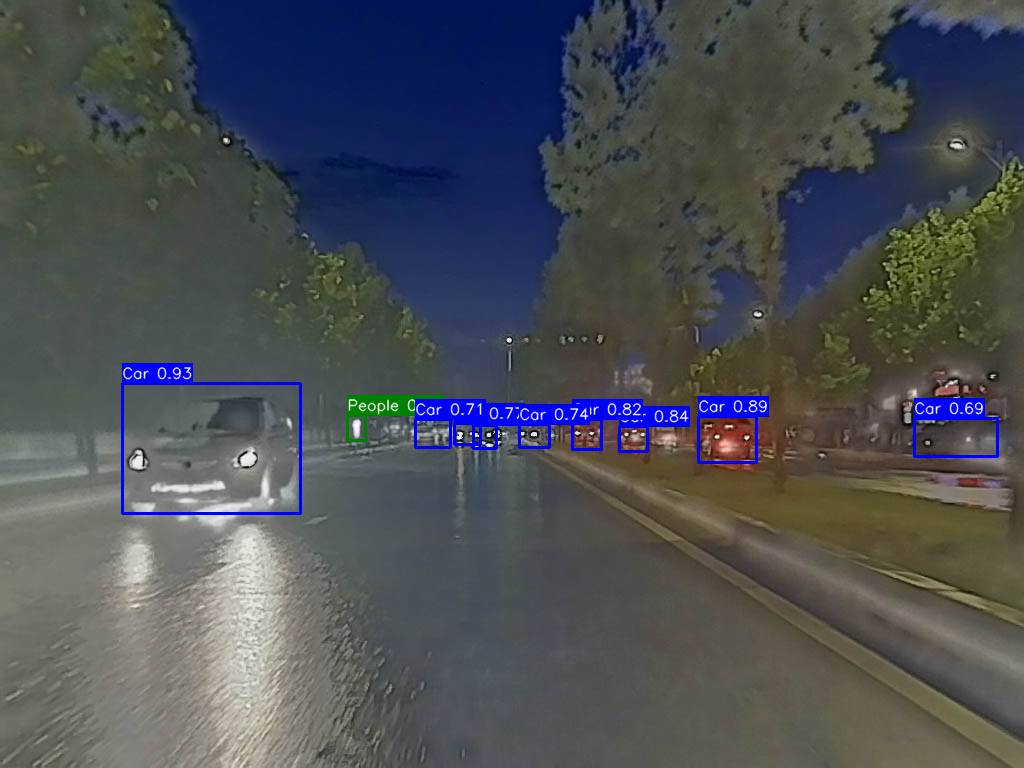}
        \caption*{Ours}
    \end{subfigure}
    \caption{Comparative visual detection of our proposed method with state-of-the-art methods on four image pairs in  \(\text{M}^{3}\text{FD}\) dataset.}
    \label{exp_detection_img}
\end{figure*}

\begin{figure}[t]
\setlength{\abovecaptionskip}{18pt} 
\setlength{\belowcaptionskip}{-15pt} 
\setlength{\intextsep}{2pt} 
    \centering
    \scriptsize
    \renewcommand\arraystretch{0.4} 
	\setlength{\tabcolsep}{0.2pt}
	\begin{tabular}{ccccccccc}	
        \multicolumn{8}{c}{\includegraphics[width=0.45\textwidth,height=0.11\textheight]{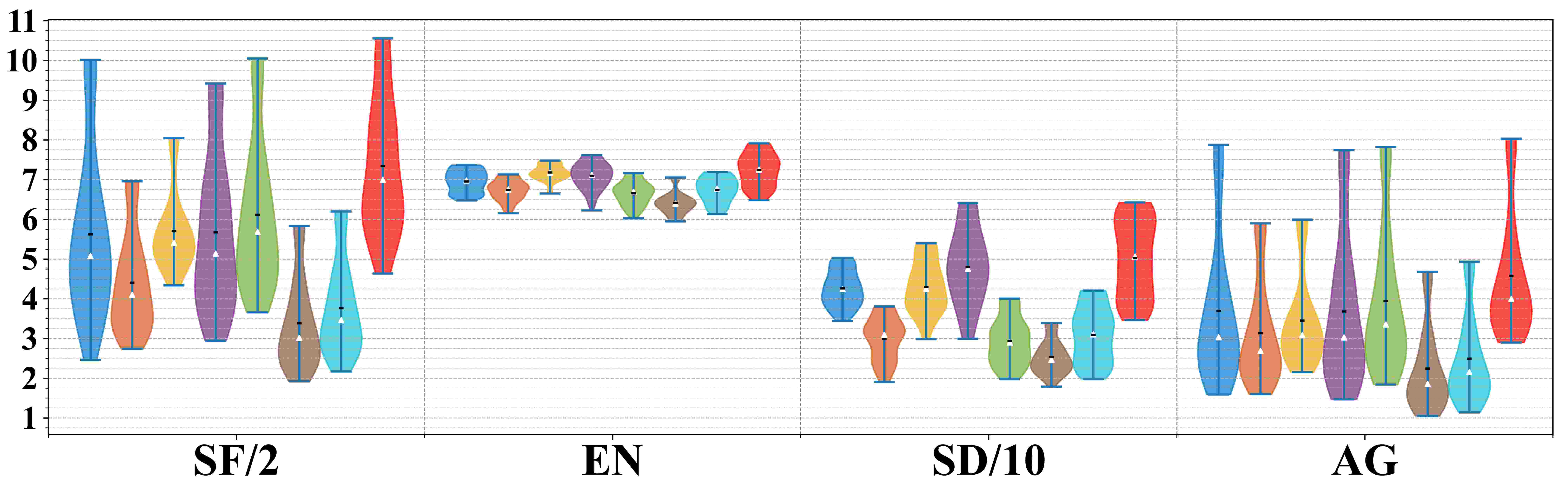}} 
        &\rotatebox{90}{\hspace{0.6cm} \textbf{M\textsuperscript{3}FD Dataset}}\\
        \multicolumn{8}{c}{\includegraphics[width=0.45\textwidth,height=0.11\textheight]{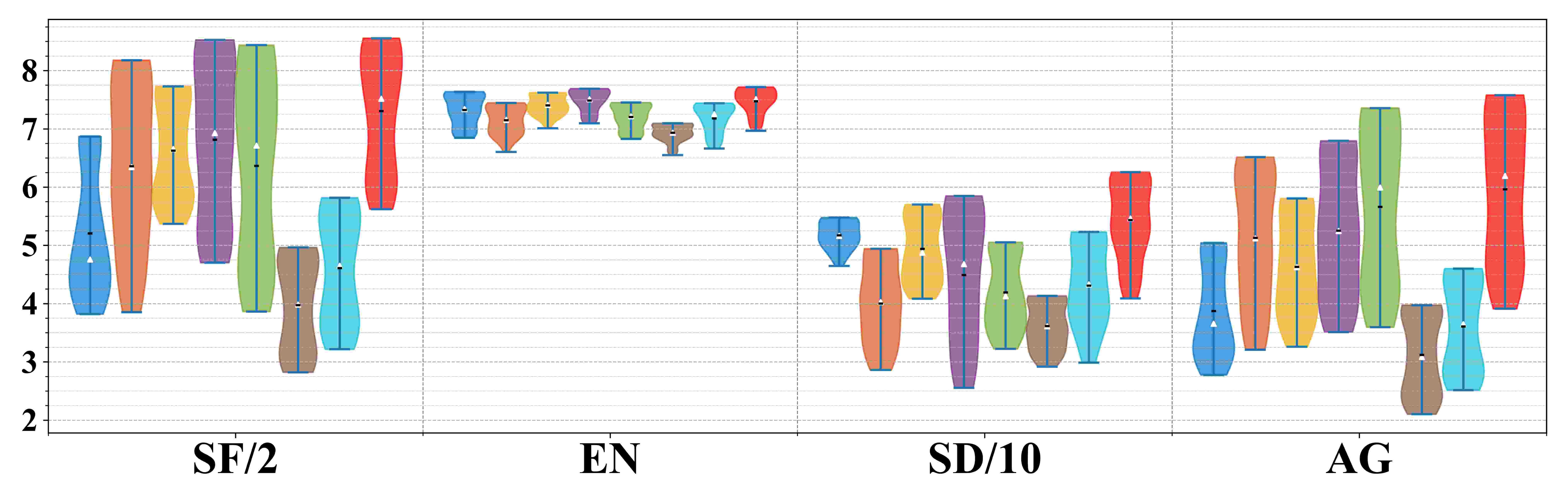}} &\rotatebox{90}{\hspace{0.3cm} \textbf{RoadScene Dataset}}\\
        \multicolumn{8}{c}{\includegraphics[width=0.45\textwidth,height=0.11\textheight]{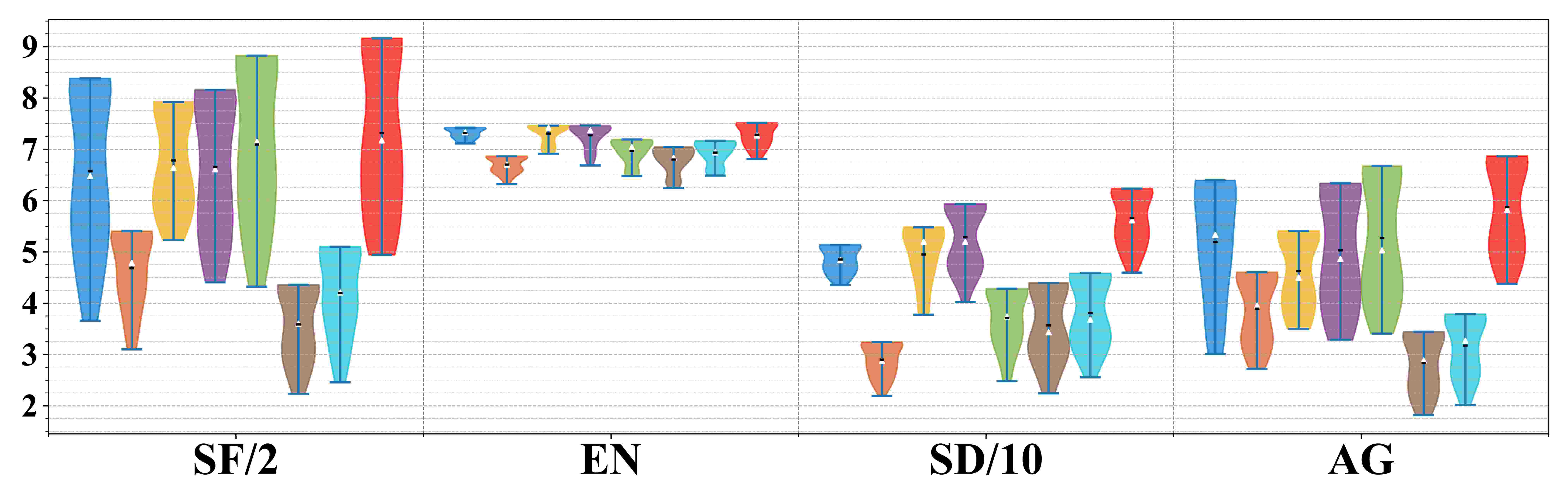}}
        &\rotatebox{90}{\hspace{0.6cm} \textbf{TNO Dataset}}\\
        \multicolumn{8}{c}{\includegraphics[width=0.45\textwidth,height=0.015\textheight]{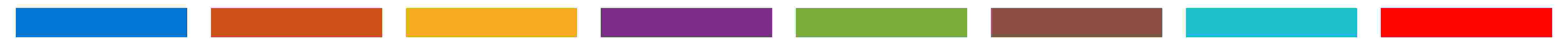}} &\\
        DDcGAN &\,U2Fusion &\,TarDAL &\;\;DDFM &\;\;CDDFuse &\;DeFusion &\;\,SegMiF &\;\,Ours
	\end{tabular}
        \vspace{-1.8em}
	\caption{Quantitative comparisons with seven IVIF methods on \(\text{M}^{3}\text{FD}\), RoadScene, and TNO datasets, respectively. The x-axis represents metrics and the y-axis are the values.}
	\label{exp_fusion_violin}
\end{figure}

\begin{figure}[h]
\setlength{\belowcaptionskip}{-10pt} 

\setlength{\intextsep}{5pt} 
    \centering
    \vspace{0em}
    \setlength{\abovecaptionskip}{2pt}
    \includegraphics[width=0.95\linewidth]{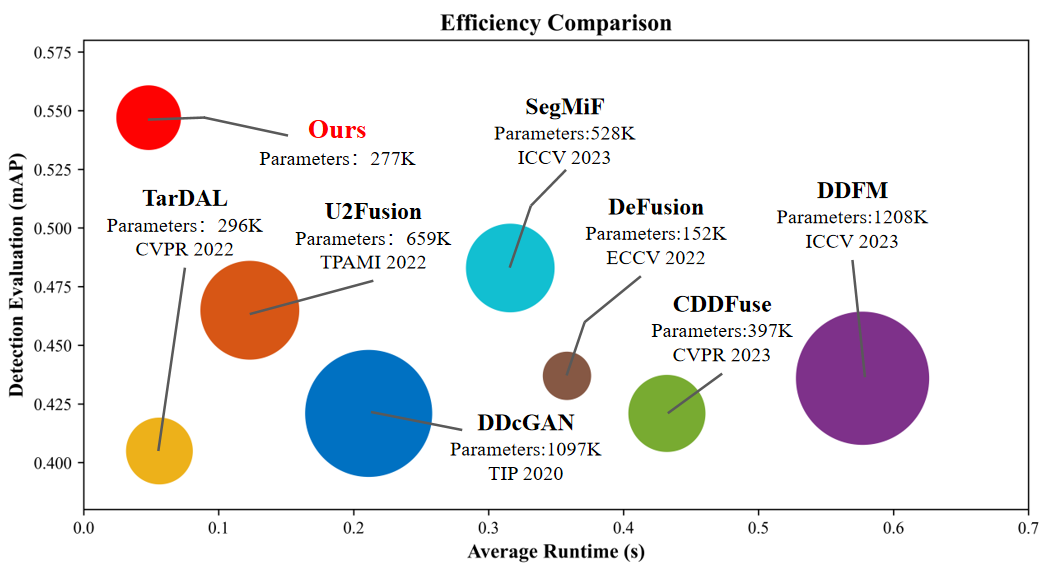}
    \caption{Comparative analysis of detection accuracy and computational efficiency against leading methods.}
    \label{exp_fusion_efficiency}
\end{figure}

\section{Experiments}
We assessed our model against several benchmarks using three publicly available datasets in the field of IVIF: \(\text{M}^{3}\text{FD}\), TNO, and RoadScene. To further validate our methodology, we generated text prompt-based versions of these datasets. Our network was trained on a GeForce RTX 3090 GPU, utilizing the Adam optimizer for parameter updates. We set the initial learning rate to \(1e^{-4}\), employing an exponential decay strategy to refine the learning process over time. The training was executed over 300 epochs with batch sizes of 64, optimizing the balance between computational efficiency and gradient precision.

\subsection{Comparative Analysis of IVIF Image Fusion}
We assess the fusion efficacy of our framework by conducting a comparative analysis with seven leading-edge methods, namely DDcGAN\cite{DDcGAN}, U2Fusion\cite{U2Fusion}, SegMiF\cite{MiFL}, DDFM\cite{DDFM}, CDDFuse\cite{CDDFuse}, DeFusion\cite{DeFusion}, and TarDAL\cite{TarDAL}. The quantitative fusion results, derived from three representative datasets, are depicted in Figure~\ref{exp_fusion_violin}. Our methodology demonstrates three principal advantages over its contemporaries. First, it proficiently conserves the thermal signatures within infrared imagery, yielding high contrast and discernibility as demonstrated in Figure~\ref{exp_fusion_img}. Second, it preserves the textural nuances of visible light images, thus resonating with the perceptual mechanisms of the human visual system, as the detection performance shown in Figure~\ref{exp_detection_img}. Third, it well amalgamates these elements, emphasizing thermal targets with greater computational efficiency achieved through the use of the codebook, as the efficiency shown in Figure~\ref{exp_fusion_efficiency}. 

As illustrated in the top row of Figure~\ref{exp_fusion_img}, our approach not only accentuates the human thermal signature but also preserves the intricate textural details and spatial resolution of the ambient environment, such as lighting and streetscape. This equilibrium is not as effectively maintained by other methods. U2Fusion, while adept at preserving texture, does not sufficiently enhance thermal targets. Both DDcGAN and DDFM are prone to introducing artifacts in their emphasis on thermal regions, potentially compromising image integrity. SegMiF and TarDAL, despite achieving a commendable balance, do not reach the level of contrast optimization that our method provides, underscoring the superiority of our approach in infrared-visible image fusion.

In the subsequent quantitative analysis, our method is rigorously benchmarked against the aforementioned state-of-the-art competitors across a comprehensive dataset comprising 397 image pairs, 37 from TNO, 60 from RoadScene, and 300 from \(\text{M}^{3}\text{FD}\). To provide a multifaceted evaluation, we employ a suite of metrics that includes Spatial Frequency (SF), Entropy (EN), Standard Deviation (SD), and Average Gradient (AG). As the quantitative results reported in Figure~\ref{exp_fusion_violin}, our method not only sets a new benchmark in maintaining high spatial frequency and average gradient but also ensures that the entropy and standard deviation of the images are superior to other state-of-the-art methods. The consistency in performance across these diverse metrics underscores the robustness and adaptability of our approach.

\subsection{Comparative Analysis of IVIF Object Detection}

As demonstrated in Figure~\ref{exp_detection_img}, our method maintains stable and precise detection capabilities even in complex environments. Obstructed vehicles and pedestrians, which pose challenges for IVIF object detection, are accurately identified through our text-guided IVIF fusion model. This robustness is attributed to the model's ability to leverage textual cues, enhancing the discriminative features that are essential for object recognition under occlusion or poor visibility conditions. The comprehensive quantitative analysis is detailed in Table~\ref{exp_detection_table}.

\begin{table}[!htbp]
\setlength{\belowcaptionskip}{-10pt} 
\setlength{\intextsep}{5pt} 
  \centering
  \small
  \renewcommand\arraystretch{1.1} 
	\setlength{\tabcolsep}{0.8mm}
    \begin{tabular}{l|ccccccc}
    \hline
    \multirow{2}{*}{Method} & \multicolumn{7}{c}{M\textsuperscript{3}FD Dataset} \\
\cline{2-8}          & Lamp  & Car   & Bus   & Motor & Truck & People & mAP \\
    \hline
    \hline
    Ir    & 0.223 & 0.711 & 0.436 & 0.331 & 0.326 & 0.507 & 0.387 \\
    Vi    & 0.243 & 0.631 & 0.462 & 0.288 & 0.274 & 0.418 & 0.359 \\
    \hline
    \cellcolor{gray!10}DDcGAN & 0.247 & 0.664 & 0.451 & 0.312 & 0.355 & 0.444 & 0.412 \\
    \cellcolor{gray!10}U2Fusion & 0.312 & \textcolor{blue}{0.724} & 0.475 & \textcolor{blue}{0.352} & 0.392 & 0.534 & 0.465 \\
    \cellcolor{gray!10}TarDAL & 0.229 & 0.652 & 0.425 & 0.285 & 0.317 & 0.525 & 0.404 \\
    \cellcolor{gray!10}DDFM  & 0.263 & 0.655 & 0.458 & 0.284 & 0.331 & 0.547 & 0.436 \\
    \cellcolor{gray!10}CDDFuse & 0.312 & 0.639 & 0.389 & 0.293 & 0.379 & 0.532 & 0.421 \\
    \cellcolor{gray!10}DeFusion & 0.324 & 0.716 & 0.469 & 0.326 & \textcolor{blue}{0.421} & 0.519 & 0.437 \\
    \cellcolor{gray!10}SegMiF  & \textcolor{blue}{0.325} & 0.722 & \textcolor{blue}{0.484} & 0.343 & 0.418 & \textcolor{blue}{0.557} & \textcolor{blue}{0.483} \\
    \hline
     \cellcolor{yellow!10}\textbf{Ours}  & \textcolor{red}{0.362} & \textcolor{red}{0.753} & \textcolor{red}{0.516} & \textcolor{red}{0.402} & \textcolor{red}{0.433} & \textcolor{red}{0.609} & \textcolor{red}{0.517} \\
    \hline
    \end{tabular}%
    \caption{Quantitative comparison of IVIF image detection on the \(\text{M}^{3}\text{FD}\) dataset. The best result is in \textcolor{red}{red} whereas the second best one is in \textcolor{blue}{blue}.}
  \label{exp_detection_table}
\end{table}%

\begin{figure}[h]
\setlength{\belowcaptionskip}{-10pt} 
	\centering
    \scriptsize
    \renewcommand\arraystretch{0.4} 
	\setlength{\tabcolsep}{0.2pt}
	\begin{tabular}{cccc}	
        \includegraphics[width=0.115\textwidth,height=0.07\textheight]{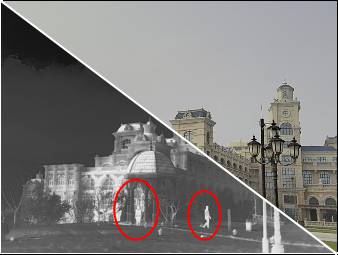} &\includegraphics[width=0.115\textwidth,height=0.07\textheight]{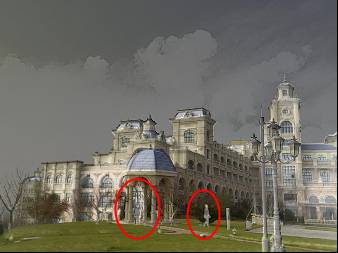}
        &\includegraphics[width=0.115\textwidth,height=0.07\textheight]{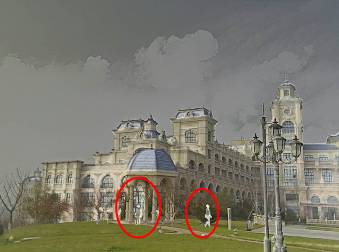}
        &\includegraphics[width=0.115\textwidth,height=0.07\textheight]{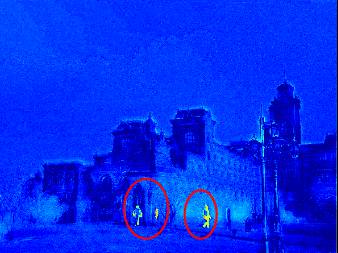}\\
        \includegraphics[width=0.115\textwidth,height=0.07\textheight]{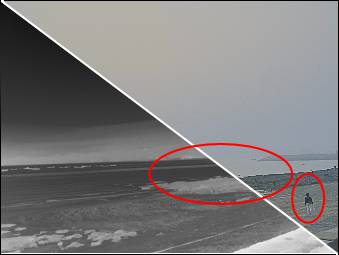} &\includegraphics[width=0.115\textwidth,height=0.07\textheight]{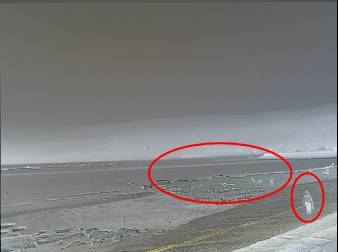}
        &\includegraphics[width=0.115\textwidth,height=0.07\textheight]{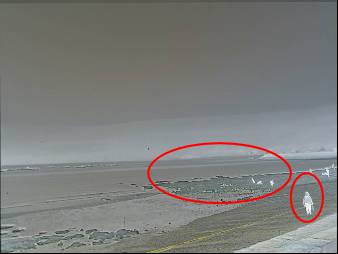}
        &\includegraphics[width=0.115\textwidth,height=0.07\textheight]{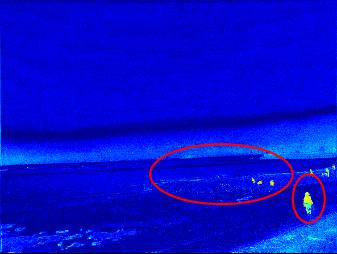}\\
        \includegraphics[width=0.115\textwidth,height=0.07\textheight]{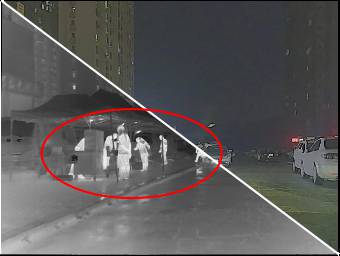} &\includegraphics[width=0.115\textwidth,height=0.07\textheight]{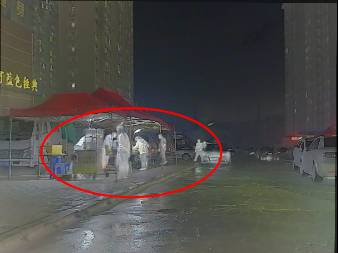}
        &\includegraphics[width=0.115\textwidth,height=0.07\textheight]{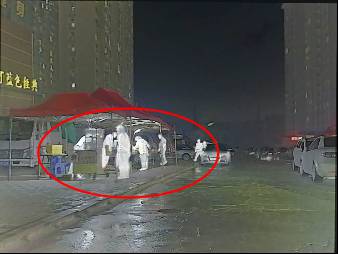}
        &\includegraphics[width=0.115\textwidth,height=0.07\textheight]{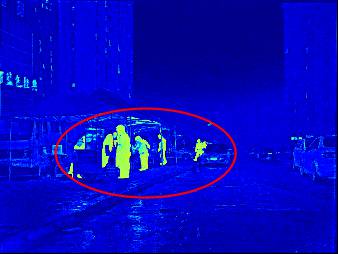}\\
        Source images & Direct &Bi-level & Difference map
	\end{tabular}\vspace{-1em}
	\caption{Visual comparison of ablation study on optimization strategies.}
	\label{ablation_optimization_img}
\end{figure}

\begin{table*}[!htbp]
\setlength{\abovecaptionskip}{3pt} 
\setlength{\belowcaptionskip}{0pt} 
\setlength{\intextsep}{5pt} 

  \centering
  \small
  \renewcommand\arraystretch{1.1} 
	\setlength{\tabcolsep}{1.3mm}
  
    \begin{tabular}{|l|cc|cccc|cccc|cccc|}
    \hline
    \multirow{2}{*}{Model} & \multicolumn{2}{c|}{Attention} & \multicolumn{4}{c|}{M\textsuperscript{3}FD Dataset} & \multicolumn{4}{c|}{RoadScene Dataset} & \multicolumn{4}{c|}{TNO Dataset} \\
\cline{2-15}          & $Att_{S}$  & $Att_{C}$  & SF    & EN    & SD    & AG    & SF    & EN    & SD    & AG    & SF    & EN    & SD    & AG \\
    \hline
    M1    &  \ding{55}     & \ding{55}      & 13.755 & 7.176 & 45.976 & 4.325 & 13.894 & 7.213 & 49.635 & \cellcolor{red!10}{5.288} & 13.318 & 7.045 & \cellcolor{red!10}{53.312} & 5.532 \\
    M2    &  \ding{51}     & \ding{55}      & 13.758 & 7.143 & 46.239 & 4.327 & 13.882 & 7.336 & 49.823 & 5.231 & 13.722 & 7.065 & 52.218 & 5.601 \\
    M3    & \ding{55}      &  \ding{51}     & \cellcolor{blue!10}{13.926} & \cellcolor{blue!10}{7.324} & \cellcolor{blue!10}{46.814} & \cellcolor{blue!10}{4.353} & \cellcolor{blue!10}{14.011} & \cellcolor{blue!10}{7.492} & \cellcolor{blue!10}{50.726} & 5.258 & \cellcolor{blue!10}{13.907} & \cellcolor{blue!10}{7.128} & 52.829 & \cellcolor{blue!10}{5.677} \\
    M4    &  \ding{51}     &  \ding{51}     & \cellcolor{red!10}{14.188} & \cellcolor{red!10}{7.413} & \cellcolor{red!10}{47.126} & \cellcolor{red!10}{4.406} & \cellcolor{red!10}{14.136} & \cellcolor{red!10}{7.512} & \cellcolor{red!10}{51.028} & \cellcolor{blue!10}{5.261} & \cellcolor{red!10}{14.103} & \cellcolor{red!10}{7.217} & \cellcolor{blue!10}{53.067} & \cellcolor{red!10}{5.689} \\
        \hline
    \end{tabular}%
    \caption{Ablation analysis on various attention mechanisms cross the \(\text{M}^{3}\text{FD}\), TNO, and RoadScene datasets. The best result is in pink whereas the second best one is in purple.}
  \label{ablation_attention_table}
\end{table*}

\begin{table}[!htbp]
\setlength{\abovecaptionskip}{3pt} 
\setlength{\belowcaptionskip}{-15pt} 
\setlength{\intextsep}{5pt} 
 \centering
  \small
  \vspace{-1em}
  \renewcommand\arraystretch{1.1} 
  \setlength{\tabcolsep}{0.8mm}
    \begin{tabular}{l|ccccccc}
    \hline
    \multirow{2}{*}{Strategy} & \multicolumn{7}{c}{M\textsuperscript{3}FD Dataset} \\
\cline{2-8}          & Lamp  & Car   & Bus   & Motor & Truck & People & mAP \\
    \hline
    \hline
    \cellcolor{gray!10}w/o cl & 0.297 & 0.641 & 0.447 & 0.336 & 0.329 & 0.528 & 0.417 \\
    \cellcolor{gray!10}w/o sl & 0.306 & 0.659 & 0.436 & 0.342 & 0.367 & 0.533 & 0.428 \\
    \hline
   \cellcolor{yellow!10}\textbf{Ours}  & \textcolor{red}{0.362} & \textcolor{red}{0.753} & \textcolor{red}{0.516} & \textcolor{red}{0.402} & \textcolor{red}{0.433} & \textcolor{red}{0.609} & \textcolor{red}{0.517} \\
    \hline
    \end{tabular}
    \caption{Quantitative ablation results of different loss functions. "cl" stands for content consistency loss, while "sl" denotes the structure loss.}
  \label{ablation_loss_table}
\end{table}

\subsection{Ablation studies}
\begin{figure}[!ht]
\setlength{\abovecaptionskip}{18pt} 
\setlength{\belowcaptionskip}{-15pt} 
\setlength{\intextsep}{5pt} 
    \centering
    \begin{subfigure}{.24\linewidth}
        \includegraphics[width=\linewidth]{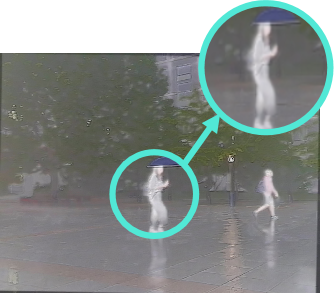}
    \end{subfigure}
    \begin{subfigure}{.24\linewidth}
        \includegraphics[width=\linewidth]{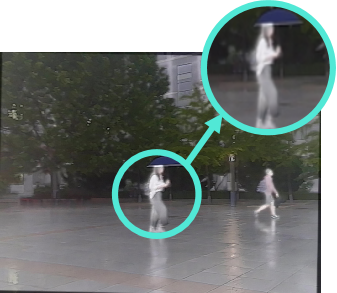}
    \end{subfigure}
    \begin{subfigure}{.24\linewidth}
        \includegraphics[width=\linewidth]{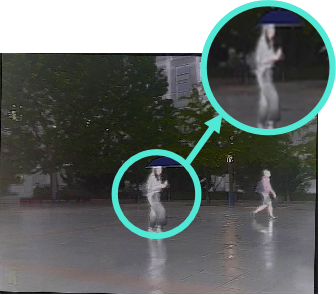}
    \end{subfigure}
    \begin{subfigure}{.24\linewidth}
        \includegraphics[width=\linewidth]{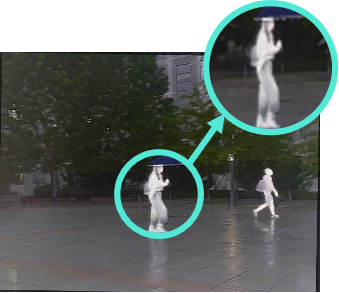}
    \end{subfigure}

    \begin{subfigure}{.24\linewidth}
        \includegraphics[width=\linewidth]{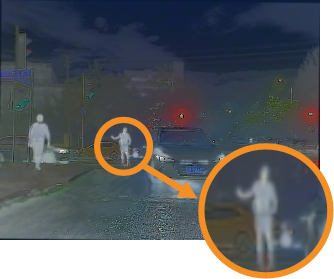}
        \caption*{base}
    \end{subfigure}
    \begin{subfigure}{.24\linewidth}
        \includegraphics[width=\linewidth]{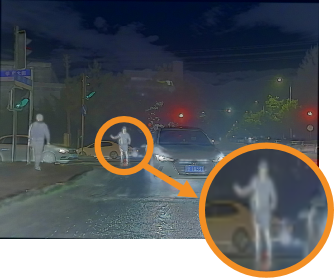}
        \caption*{w/o \(Att_{C}\)}
    \end{subfigure}
    \begin{subfigure}{.24\linewidth}
        \includegraphics[width=\linewidth]{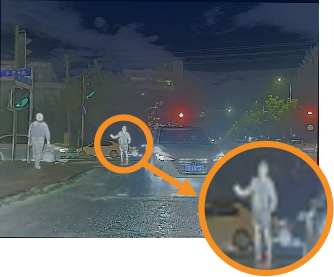}
        \caption*{w/o \(Att_{S}\)}
    \end{subfigure}
    \begin{subfigure}{.24\linewidth}
        \includegraphics[width=\linewidth]{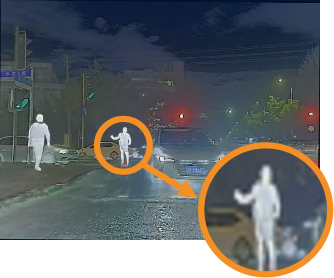}
        \caption*{full}
    \end{subfigure}
   
    \caption{Visual comparison of various attention mechanisms. \(Att_{C}\) stands for cross-attention, while \(Att_{S}\) denotes the self-attention.}
    \label{ablation_attention_img}
    
\end{figure}

\paragraph{Evaluating different attention mechanisms}
As the results shown in Table~\ref{ablation_attention_table}, we assess the impact of self-attention and cross-attention mechanisms on our network for target detection. Four scenarios were tested: both self-attention and cross-attention, only cross-attention, only self-attention, and neither. Visual results shown in Figure~\ref{ablation_attention_img} indicate that using both mechanisms achieves the best performance, highlighting their complementary roles in enhancing image features and integrating these with textual information. The absence of one or both mechanisms notably reduces the model's effectiveness, emphasizing the importance of these attention processes in our model.

\begin{figure}[!h]
\setlength{\abovecaptionskip}{15pt} 
\setlength{\belowcaptionskip}{-16pt} 
\setlength{\intextsep}{5pt} 
    \centering
    \begin{subfigure}{.19\linewidth}
        \includegraphics[width=\linewidth]{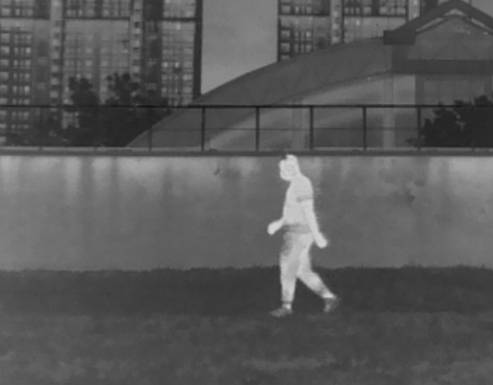}
    \end{subfigure}
    \begin{subfigure}{.19\linewidth}
        \includegraphics[width=\linewidth]{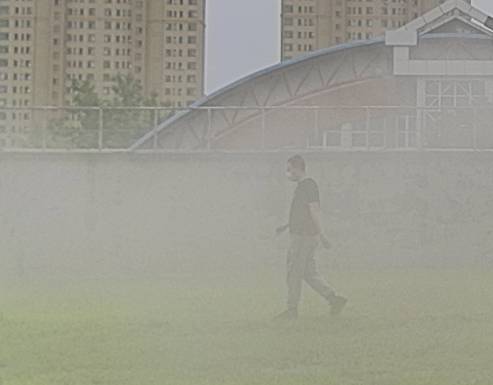}
    \end{subfigure}
    \begin{subfigure}{.19\linewidth}
        \includegraphics[width=\linewidth]{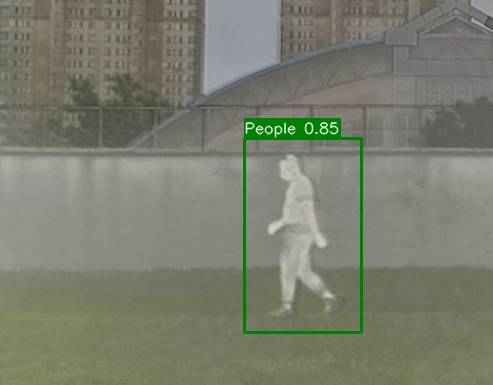}
    \end{subfigure}
    \begin{subfigure}{.19\linewidth}
        \includegraphics[width=\linewidth]{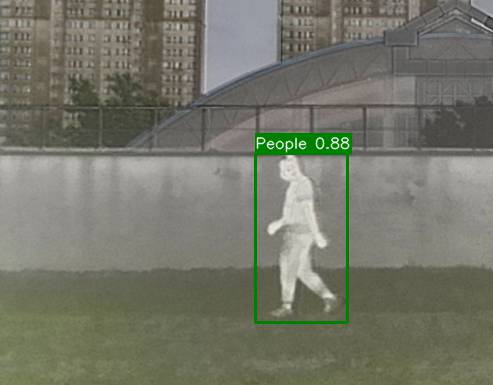}
    \end{subfigure}
    \begin{subfigure}{.19\linewidth}
        \includegraphics[width=\linewidth]{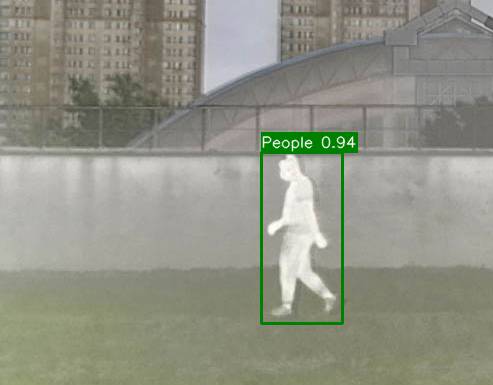}
    \end{subfigure}

    \begin{subfigure}{.19\linewidth}
        \includegraphics[width=\linewidth]{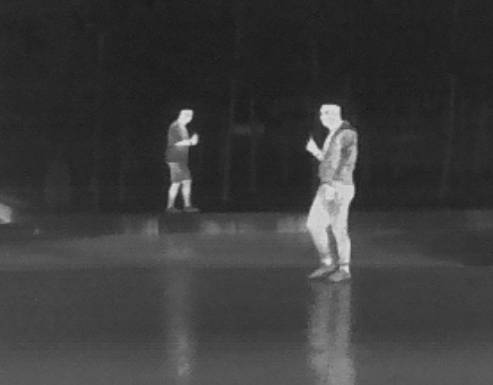}
        \caption*{Ir}
    \end{subfigure}
    \begin{subfigure}{.19\linewidth}
        \includegraphics[width=\linewidth]{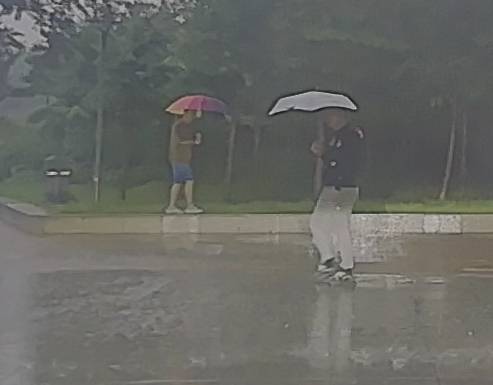}
        \caption*{Vi}
    \end{subfigure}
    \begin{subfigure}{.19\linewidth}
        \includegraphics[width=\linewidth]{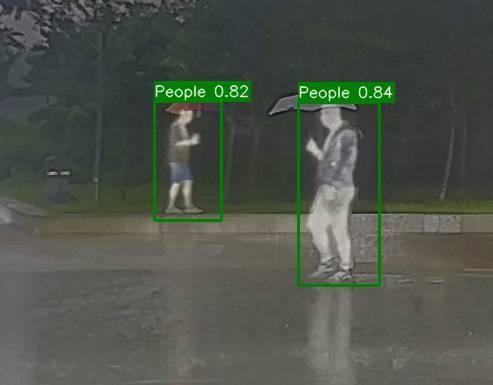}
        \caption*{w/o cl}
    \end{subfigure}
    \begin{subfigure}{.19\linewidth}
        \includegraphics[width=\linewidth]{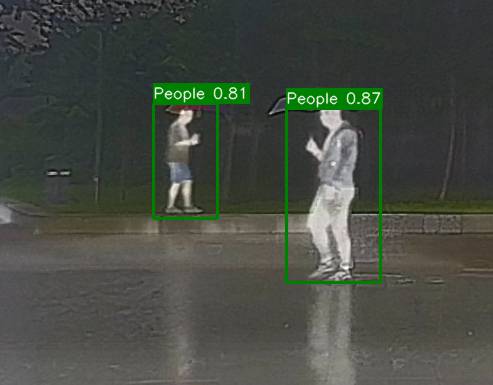}
        \caption*{w/o sl}
    \end{subfigure}
    \begin{subfigure}{.19\linewidth}
        \includegraphics[width=\linewidth]{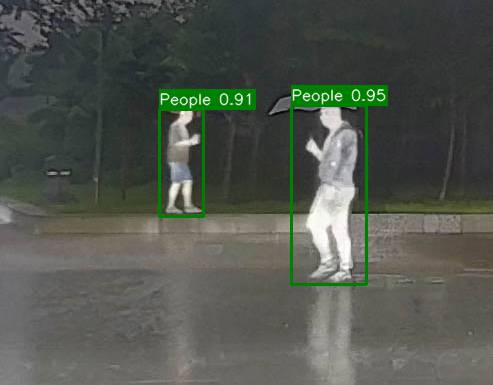}
        \caption*{full}
    \end{subfigure}
   
    \caption{Visual comparison of ablation study on loss function. "sl" denotes the structure loss and "cl" represents the content consistency loss.}
    \label{ablation_loss_img}
\end{figure}

\paragraph{Analyzing the training loss functions}
The impact of structure loss and content consistency loss is discussed in Figure~\ref{ablation_loss_img}. Structure loss ensures the stability of the model across epochs, while content consistency loss maintains the alignment with the inputs. By training models without one or both of these components, we observed a significant degradation in fusion and detection performance, as illustrated in our quantitative results Table~\ref{ablation_loss_table}. This degradation highlights the critical role of both structure loss and content consistency loss in maintaining the quality and coherence of the fused output, confirming their indispensable contribution to the effectiveness of our target detection model.

\paragraph{Experiments on training strategy}
Figure~\ref{ablation_optimization_img} showcases the enhancements achieved using our bilevel optimization approach relative to the direct joint training with our proposed network. Our strategy not only facilitates the the superior image fusion quality but also sustains the discernibility and fidelity, under severe conditions. This obtains a significant advantage in enhancing the detection performance and improving visual effects.


\begin{figure}[!ht]
\setlength{\abovecaptionskip}{15pt} 
\setlength{\belowcaptionskip}{-15pt} 

\setlength{\intextsep}{5pt} 
    \begin{subfigure}{.56\linewidth}
        \includegraphics[width=\linewidth]{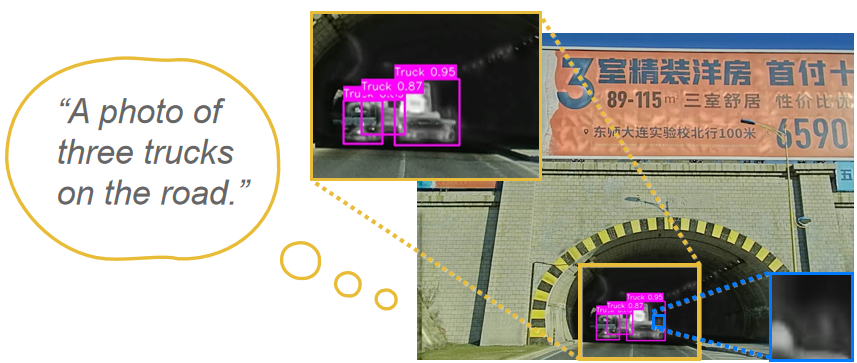}
        \caption*{}
    \end{subfigure}
    \begin{subfigure}{.34\linewidth}
        \includegraphics[width=\linewidth]{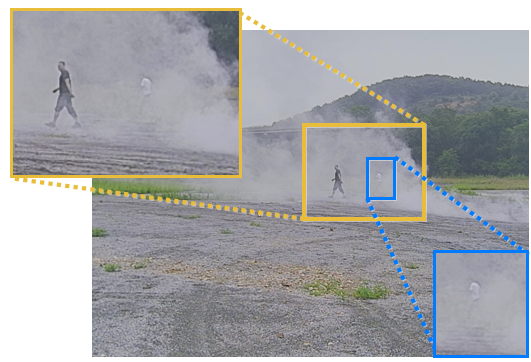}
        \caption*{}
    \end{subfigure}
    
    \begin{subfigure}{.56\linewidth}
        \includegraphics[width=\linewidth]{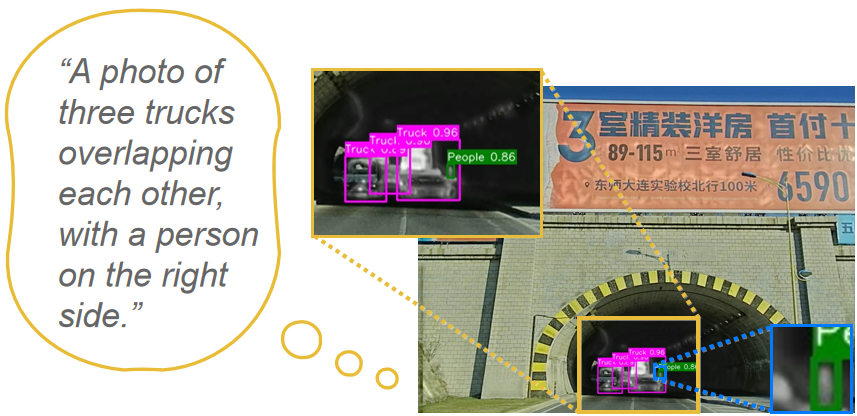}
        \caption*{detection}
    \end{subfigure}
    \begin{subfigure}{.34\linewidth}
        \includegraphics[width=\linewidth]{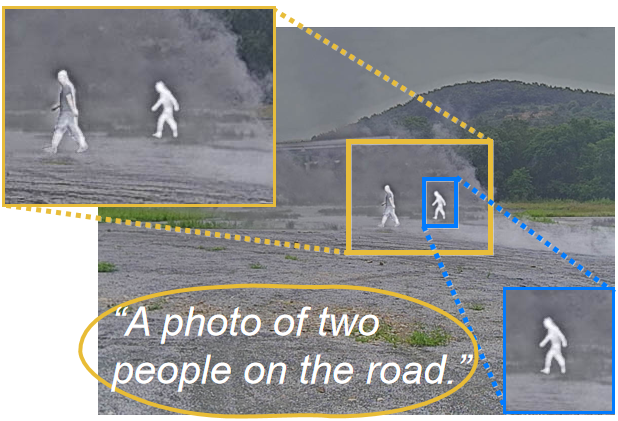}
        \caption*{fusion}
    \end{subfigure}

    \caption{Visual effects of textual prompts on detection and fusion. The left column delineates the influence exerted by both coarse and refined prompts on detection. The right column presents an analysis on the fusion, showing results with and without the textual prompts.}
    \label{ablation_prompt}
    
\end{figure}

\paragraph{Impact of different text prompts}
The ablation study visualized in Figure~\ref{ablation_prompt} examines the impact of textual prompts on the fusion efficacy and detection performance within our proposed network. The right series of images delineates the perceptual distinctions in fusion when textual prompts are introduced versus their absence. Notably, the integration of textual prompts enhances the image's brightness and accentuates key features, confirming the prompts' pivotal role in directing the fusion process toward more pronounced elements. The left column offers a more granular analysis, contrasting detection results between a coarse text prompt and a fine text prompt. The additional text prompt information enables our model to surpass ground-truth annotations in terms of model comprehension, as evidenced by the enhanced detection of previously unmarked subjects. This qualitative enhancement validates the text prompts' effectiveness in guiding the network to focus on and amplify the most critical aspects of the imagery.

\section{Conclusion}
We propose the first text-guided multi-modality image fusion network, specifically for object detection, leveraging the CLIP model to bridge the semantic gap between infrared and visible imagery. Our approach, featuring a bilevel optimization strategy and the utilization of a codebook, not only enhances the alignment of text and image features but also significantly improves the detection accuracy and efficiency. The creation of the first dataset with paired infrared and visible images and accompanying text prompts sets a precedent in this research domain.

{
    \small
    \bibliographystyle{ieeenat_fullname}
    \bibliography{main}
}

\end{document}